\documentclass[runningheads]{llncs}
\usepackage{graphicx}

\usepackage{tikz}
\usepackage{comment}
\usepackage{amsmath,amssymb} 
\usepackage{color}
\usepackage{hyperref}
\usepackage{algorithm}
\usepackage[noend]{algpseudocode}
\usepackage{paralist}
\usepackage[lofdepth,lotdepth]{subfig}
\usepackage{array,multirow}

\usepackage{url}
\newcommand{\cQ}{\mathcal{Q}}

\newcommand{\cD}{\mathcal{D}}

\newcommand{\bE}{\mathbf{E}}

\newcommand{\cS}{\mathcal{S}}

\newcommand{\bO}{\mathbf{O}}
\newcommand{\bo}{\mathbf{o}}

\newcommand{\bp}{\mathbf{p}}

\newcommand{\bbR}{\mathbb{R}}

\newcommand{\bv}{\mathbf{v}}

\newcommand{\cW}{\mathcal{W}}

\newcommand{\cV}{\mathcal{V}}

\newcommand{\cH}{\mathcal{H}}

\begin{document}

\pagestyle{headings}
\mainmatter

\title{Visual Localization Under Appearance Change: Filtering Approaches} 


\institute{School of Computer Science, The University of Adelaide}



\titlerunning{Visual Localization Under Appearance Change: Filtering Approaches}
%
\author{Anh-Dzung Doan  \and Yasir Latif \and  Tat-Jun Chin \and Yu Liu \and Shin-Fang Ch’ng \and Thanh-Toan Do \and Ian Reid}
\authorrunning{Anh-Dzung Doan, et al.}
%
\institute{School of Computer Science, The University of Adelaide, Australia}

\maketitle

\begin{abstract}
A major focus of current research on place recognition is visual localization for autonomous driving. In this scenario, as cameras will be operating continuously, it is realistic to expect videos as an input to visual localization algorithms, as opposed to the single-image querying approach used in other visual localization works.  In this paper, we show that exploiting temporal continuity in the testing sequence significantly improves visual localization - qualitatively and quantitatively. Although intuitive, this idea has not been fully explored in recent works. To this end, we propose two filtering approaches to exploit the temporal smoothness of image sequences: i) filtering on discrete domain with Hidden Markov Model, and ii) filtering on continuous domain with Monte Carlo-based visual localization. Our approaches rely on local features with an encoding technique to represent an image as a single vector. The experimental results on synthetic and real datasets show that our proposed methods achieve better results than state of the art (i.e., deep learning-based pose regression approaches) for the task on visual localization under significant appearance change. Our synthetic dataset and source code are made publicly available\footnote{https://sites.google.com/view/g2d-software/home}\footnote{https://github.com/dadung/Visual-Localization-Filtering}.

\end{abstract}

\section{Introduction}

To carry out higher level tasks such as planning and navigation, 
a robot needs to maintain, at all times, an accurate estimate of its position and orientation with respect to the environment. When the robot uses an existing map to infer its 6 degree of freedom (DoF) pose, the problem is termed \textit{localization}. In the case when the map information is appearance (images) associated with different parts of the map, the problem is that of \textit{visual localization} (VL).
Image-based localization methods normally assume that the appearance remains unchanged from when the map is generated to the present time when the robot needs to localize itself. 
However, as the operational time span of the robot increases, the appearance of the environment inevitably changes.
This poses a great challenge for visual localization methods as the underlying assumption of static appearance is violated due to continual changes of environment, e.g., weather condition, time of day, construction sites, updating of fa\c{c}ades and billboards, etc.

One approach towards dealing with appearance change is to observe as many variations as possible of each location and carry a representation that can model them \cite{churchill2013experience,doan2019scalable}. However, the bottleneck in this case is the significant amount of time needed to capture sufficient naturally occurring appearance variations.
For example, recent benchmarks \cite{sattler2018benchmarking,apolloscape,OxfordRobotCar} have taken years to collect data with sufficient natural appearance variations.

An orthogonal approach to address appearance change, pioneered by SeqSLAM \cite{milford2012seqslam}, is to consider sequence-to-sequence matching instead of matching a single query image to previously observed images.
SeqSLAM showed that sequence-based matching is more resilient to appearance changes but comes with its own shortcoming (e.g., sensitivity to view-point changes and differences in sequence velocities~\cite{sunderhauf2013we}). 
While recent learning-based VL algorithms \cite{brachmann2017dsac,kendall2015posenet,sattler2017efficient,kendall2017geometric} have focused on the case of a single query image, the camera, in robotics setting, is operating continuously once the vehicle is moving and hence, it is realistic to expect a video (sequence of images) as an input to VL methods. 

Another paradigm to overcome this challenge is to extract ``fingerprints" which are robust against appearance change \cite{torii2015densevlad,arandjelovic2016netvlad}. However, this approach has been only shown its efficiency in ``natural" changes, e.g., weather conditions, or times of day. 

\noindent \textbf{Our contributions:} This work addresses the problem of metric visual localization (i.e., we recover the 6 DoF pose of the camera) under appearance change, in the context of autonomous vehicles. To effectively reason over an image sequence (video), we propose two visual localization (VL) methods:

\begin{itemize}
	\item  Visual localization with Hidden Markov Model (HMM): the inference is performed on image indices to retrieve place hypotheses from which query frames are possibly captured. The 6 DoF camera poses of query frames are then interpolated from those place hypotheses. The inference is handled through a transition matrix to effectively exploit the temporal continuity of query sequences.
	\item Monte Carlo-based visual localization: the Monte Carlo principle is leveraged to approximate the probability distribution of 6 DoF camera poses incrementally. 
	We show that for the case of driving on urban streets, a simple motion model is enough to successfully track the vehicle's pose. 
\end{itemize}

The above approaches rely on a novel \textit{observation encoder} which generates a fixed (low) dimensional vector representation for each image. We show experimentally that such representation is more resilient to appearance change along with minor changes in the viewpoint, while being compact in size.

We validate the proposed localization methods on synthetic and real datasets \cite{OxfordRobotCar}  to show that they outperform state-of-the-art methods (i.e., deep learning-based camera pose regression) in the task of 6 DoF visual localization in the large-scale environment with significant appearance change, i.e., the traversal distance of the vehicle is about 10 km.

\section{Related Works}
Visual localization is a very well-studied area and various methods have been proposed to address the problem in different ways. For the scope of our work, we limit the survey to methods that can perform metric localization, i.e., they recover the 6 DoF pose of the camera based on image information.
These algorithms can be categorized into local feature-based methods and learning-based methods.

Broadly speaking, local feature-based methods estimate the pose of the query image from 2D-3D correspondences by solving the PnP problem \cite{lepetit2009epnp} or an equivalent step \cite{sattler2018benchmarking}. The main difficulty for this approach is establishing and validating the 2D-3D correspondences for large (city-scale) maps. 
Many methods attempt to improve this step by using 
geometric constraints \cite{sattler2017efficient,tran2019device}, semantic information in the images \cite{schonberger2018semantic}, and 
learning-based matching \cite{brachmann2017dsac}.

Learning-based methods perform 6 DoF pose estimation as an image-to-SE(3) regression problem. PoseNet \cite{kendall2015posenet} uses a convolutional neural network (CNN) to learn the regression mapping. 
LSTM-Pose \cite{walch2017lstm} argues that the usage of a fully connected layer in PoseNet possibly leads to the overfitting, despite the usage of dropout. Therefore, they use Long-Short Term Memory (LSTM) units. 
Other variants of PoseNet include: uncertainty in the pose prediction~\cite{kendall2016modelling} and geometric constraints~\cite{kendall2017geometric,brahmbhatt2018mapnet}.

Our proposed method is different from above methods, 
in that we seek not only the current 6 DoF pose of the camera as a point estimate but as a distribution over the space of all possible locations in the map. Therefore, we employ probabilistic filtering frameworks, namely Hidden Markov Model and Monte Carlo localization, that take into account the constraints on motion of the vehicle and the temporal nature of the localization task.
To deal with appearance changes, 
we propose a novel observation encoder which relies on dense local features  and an encoding technique to ensure the robustness against environmental changes. While, previous works \cite{wolf2005robust,wolf2002robust,menegatti2004image} used probabilistic filtering-based methods for localizing a robot in an indoor environment, they only consider the case of a robot moving on plane (3 DoF) while we aim to recover the complete 6 DoF pose of the robot. 
Also, while these works \cite{wolf2005robust,wolf2002robust,menegatti2004image} only test their methods in the indoor environment, in which the appearance change is insignificant; we show that our methods can perform robustly in a large-scale outdoor environment with significant appearance change, i.e., the car traverses over 10 km in its run.

\section{Problem definition}
We define the map $\cD = \{ \cV^1, ..., \cV^n \}$, where $\cV^i = \{ (I_1^i, s_1^i), ..., (I_T^i, s_T^i) \} $ is an image sequence (video) $i$. Every frame $I^i_t$ is associated with a 6 DoF camera pose $s^i_t$, which can be attained by conducting Visual SLAM \cite{parra2019visual} or Structure from Motion \cite{schonberger2016structure}.

For the online inference, given a query video $\cQ = \{ Q_1, ..., Q_T \}$, our goal is to estimate the camera pose $s_t$ for each $Q_t$ with respect to the map $\cD$.

As our both methods rely on the observation encoder which maps every image to a single vector, we are going to describe it first in section \ref{sec:obs_enc}. Subsequently, our proposed visual localization using HMM and Monte Carlo principle are respectively detailed in sections \ref{sec:vl_hmm} and \ref{sec:MCVL}. After that, section \ref{sec:g2d_and_dataset} presents our synthetic dataset collected from the computer game Grand Theft Auto V (GTA V) using G2D \cite{Dzung18G2D}. Finally, experimental results and discussion are reported in section \ref{sec:experiments} and \ref{sec:discussion}.

\section{Observation encoder}
\label{sec:obs_enc}

We seek for every image $I$ a nonlinear function $\tau(I)$ that maps $I$ to a vector in a fixed dimensional space. To do so,
we first densely compute SIFT features: $\{x_i \in \mathbb{R}^d \, | \, i = 1, ...,m \}$ over the image, followed by RootSIFT normalization \cite{arandjelovic2012three}:

\begin{inparaenum}[i)] 
	\item $L$1 normalize every SIFT vector $x_i = x_i / ||x_i||_1$ 
	
	\item square root each element $x_i = \sqrt{x_i}$ 
\end{inparaenum}. 

\noindent where, RootSIFT normalization makes the Euclidean distance calculation among SIFT features equivalent to computing the Hellinger kernel.    

Subsequently, we employ the embedding function VLAD \cite{jegou2010vlad} to embed SIFT features into a higher dimensional vector space. 
In particular, given a vocabulary learned by K-means: $\mathcal{C} = \{c_k \in \mathbb{R}^d \, | \, i = 1, ...,K \}$, every SIFT feature is embedded as follows: 

\[\phi_{VLAD}(x_i) = [..., 0, x_i - c_j, 0, ...] \in \mathbb{R}^D\]

\noindent where $c_j$ is the nearest visual word to $x_i$, and $D = K \times d$. Note that different from Bag of Word (BoW), which embeds the feature vector as follows: 

\[\phi_{BoW}(x_i) = [..., 0, 1, 0, ...] \in \mathbb{R}^K\]

\noindent where only $j^{th}$ component of $\phi_{BoW}(x_i)$ non-zero means that the nearest neighbor of feature $x_i$ is visual word $c_j$; VLAD considers the residual between a visual word and its nearest feature. Do et al. \cite{do2015faemb} show that VLAD is a simplified version of local coordinate coding \cite{yu2010improved}, which tries to approximate a nonlinear classification function by a linear function.

From a set of embedded vector: $\{\phi(x_i) \in \mathbb{R}^D \, | \, i = 1, ...,m \}$, we aggregate them by the sum pooling to obtain a single vector representing the image $I$: 

\[\tau(I) = \displaystyle\sum_{i=1}^{m} \phi(x_i)\]

In literature, there are several other ways for this aggregation step \cite{jegou2014triangulation,murray2014generalized}, but for simplification, we choose sum pooling and show that it can obtain good performance in practice.

One drawback of using local features in image retrieval is that the background features (e.g., trees, sky, road, etc) significantly outnumber features from informative objects (e.g., buildings). 
To alleviate this, we apply PCA projection, whitening and L2-normalization \cite{jegou2012pcawhitening}, which limit the impact of background features in the vector $\tau(I)$.

\section{Visual Localization with HMM}
\label{sec:vl_hmm}

\begin{figure}
	\centering
	\includegraphics[width=1\textwidth]{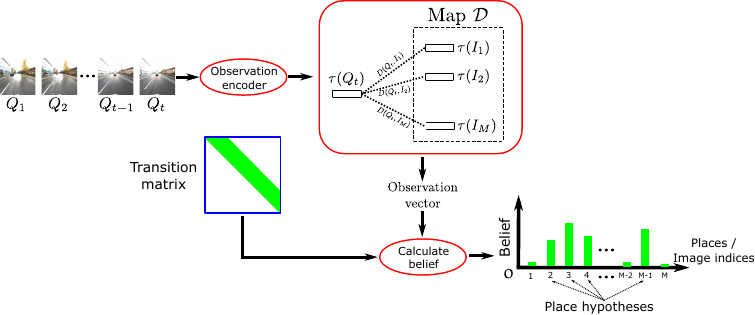}
	\caption{The diagram of our proposed HMM for visual localization. Assume the map $\cD$ has $M$ images after ``unrolling" all videos $\cV^i = \{ (I_1^i, s_1^i), ..., (I_T^i, s_T^i) \}$, where each image indices is regarded as a physical place. Places with top highest belief probabilities are selected as place hypotheses, which are then used for estimating 6 DoF pose (see section \ref{sec:vl_hmm:est_pose}).}
	\label{fig:hmm_diagram}
\end{figure}

Let the map $\cD = \{I_j, s_j\}_{j=1}^M$ be the database of $M$ images after ``unrolling" all videos $\cV^i = \{ (I_1^i, s_1^i), ..., (I_T^i, s_T^i) \}$. We regard each image index $j$ as a \textit{``place"}. If the field of view of two images are overlap sufficiently (e.g., if they are temporally close in their source video), we will assign a transition probability between these two image indices. The detail of computing a full transition matrix of $\cD$ will be given in section \ref{sec:vl_hmm:trans_matrix}.
	
Given a query video $\cQ = \{ Q_1, ..., Q_T \}$, our goal is to find place hypotheses $\cH_t$ for every $Q_t$ (see section \ref{sec:vl_hmm:place_hypotheses}). Then, the 6 DoF camera pose of every $Q_t$ will be estimated through place hypotheses $\cH_t$ (see section \ref{sec:vl_hmm:est_pose})

\subsection{Place hypotheses from HMM} \label{sec:vl_hmm:place_hypotheses}

To find the place hypotheses on query video $\cQ$, we model $\cQ$ using a Hidden Markov Model (HMM). We regard each image $Q_t$ as a noisy measurement of \textit{place state $v_t$}, where $v_t \in \{j\}_{j=1}^M$, and as mentioned above, $\{j\}_{j=1}^M$ is set of image indices (or places) of database $\cD$

A Hidden Markov Model (HMM) consists of three parameters $\{\bE, \bO, \pi\}$, which respectively are transition matrix, observation model and initial state probabilities. Size of HMM parameters are: $\bE \in \bbR^{M \times M}$, $\bO \in \bbR^{M \times T}$ and $\pi \in \bbR^M$. Specifically, 

\[
\bE(j_1, j_2) = P(v_t = j_2 | v_{t-1} = j_1)
\]

\noindent where, the value at row $j_1$ and column $j_2$ of matrix $\bE$ represents the probability of transitioning from place $j_1$ to place $j_2$.

The observation matrix is defined as follows:

\[
\bO(j, t) = P(Q_t | v_t = j)
\]

\noindent which represents the likelihood probability of the frame $I_j$ matched against query frame $Q_t$

The recursive belief $\bp_t = P(v_t | Q_{1:t})$ is computed in the following:

\begin{equation}
	\bp_t  = \eta \: \bo_t \circ \bE^T \bp_{t-1}
	\label{eq:belief}
\end{equation}

\noindent where, $\bp_t(j)$ is the belief probability of place $j$, given queries up to time $t$ (i.e., $Q_{1:t}$). $\circ$ is Hadamard product, $\bo_t$ is the $t^{th}$ column of $\bO$, $\eta$ is the normalizing constant to ensure $\sum \bp_t = 1$. At time $t=1$, $\bp_1 = \pi$, where $\pi(j) = 1/M$.

The place hypotheses $\cH_t$ at time $t$ are top $k$ places with largest $\bp_t(j)$:

\begin{equation}
	\cH_t = \{j \; | \; \text{top $k$ largest} \;\bp_t(j) \}
	\label{eq:place_hypotheses}
\end{equation}

\subsubsection{Transition matrix}\label{sec:vl_hmm:trans_matrix}
Given an video $\cV^i = \{I_1^i, ..., I_T^i \}$, a transition matrix $E^i$ of size $T \times T$ is formed as the following:

\[
E^i(r, c) = \begin{cases}
	1 & 0 \le c-r \le \bv_{\text{max}} \\
	0 & \text{otherwise}
\end{cases}
\]

\noindent where, $E^i(r, c)$ is the element at row $r^{\text{th}}$ and column $c^{\text{th}}$. $\bv_{\text{max}}$ is the maximum velocity of the vehicle. The transition matrix $\bE$ of database $\cD$ is then created by concatenating every $E^i$ in the diagonal direction:

\[
\bE = \begin{bmatrix}
	E^1 &  &  & \\
	& E^2 &   & \\
	&  & \ddots & \\
	& & & E^n \\
\end{bmatrix}
\]

Subsequently, transition matrix $\bE$ is  normalized to ensure the summation of every row of $\bE$ equals to $1$, i.e., $\displaystyle\sum_{c} \bE(r, c) = 1$

\subsubsection{Observation model}

The distance between $Q_t$ to every database image $I_j \in \cD$ is computed in the following

\begin{equation}
	D(Q_t, I_j) = \begin{Vmatrix} \tau(Q_t) - \tau(I_j)\end{Vmatrix}_2^2
	\label{eq:square_distance}
\end{equation}

\noindent where, $\tau(.)$ is the observation encoder (see Section \ref{sec:obs_enc}). The observation vector $\bo_t$ is then calculated:

\begin{equation}
	\bo_t = \text{exp}\begin{pmatrix}-\frac{D}{\sigma}\end{pmatrix}
	\label{eq:observation_vector}
\end{equation}

\subsection{Estimating 6 DoF pose}
\label{sec:vl_hmm:est_pose}
Due to every image in $\cD$ associated with a corresponding 6 DoF camera pose, mean-shift algorithm is employed on $\cH_t$ over the translational part of their poses. The largest cluster is then selected to calculate the mean of translation and rotation \cite{markley2007averaging}, which is regarded as predicted 6 DoF pose $s_t$ of the image query $Q_t$.

\subsection{Overall algorithm}
The Algorithm \ref{alg:place_hypotheses} presents the method of visual localization with HMM. Given the query $Q_t$, we calculate its fixed high-dimensional vector representation $\tau(Q_t)$. The Euclidean distances between $Q_t$ and every database frames $I_j$ is computed, which is then used to estimate the observation vector $\bo_t$. The belief $\bp_t$ is subsequently computed which is utilized to form place hypotheses $\cH_t$. The predicted pose of $Q_t$ is finally estimated from place hypotheses $\cH_t$.

\begin{algorithm}
	\caption{Visual localization with HMM}
	\begin{algorithmic}[1]
		\Function{HMM}{$Q_t$, $\cD$, $\bE$}
		\State Compute $\tau(Q_t)$ as Section \ref{sec:obs_enc}
		\State Compute $D\begin{pmatrix}Q_t, I_j\end{pmatrix}$, $\forall I_j \in \cD$ using equation (\ref{eq:square_distance}).
		\State Compute observation vector $\bo_t$ using equation (\ref{eq:observation_vector})
		\State Compute belief $\bp_t$ using equation (\ref{eq:belief})
		\State Form place hypotheses $\cH_t$ using equation (\ref{eq:place_hypotheses})
		\State Estimate 6 DoF pose $s_t$ of $Q_t$ (section \ref{sec:vl_hmm:est_pose})
		\State \textbf{return} $s_t$
		\EndFunction
	\end{algorithmic}
	\label{alg:place_hypotheses}
\end{algorithm}

\section{Monte Carlo-based visual localization (MCVL)}
\label{sec:MCVL}

\begin{figure}
	\centering
	\includegraphics[width=0.6\textheight]{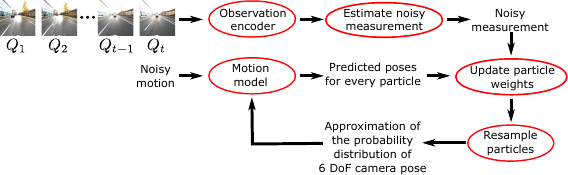}
	\caption{The diagram of our proposed MCVL.}
	\label{fig:pf_diagram}
	\vspace{-1em}
\end{figure}

Let the 6 DoF camera pose of $Q_t$ be given by: 

\[s_t = [r_t, \, \Omega_t]^T\]

\noindent where, $r_t$ and $\Omega_t$ represent the 3D position and Euler orientation respectively at time $t$. Given motion $u_{1:t}$ and noisy measurement $z_{1:t}$ up to time $t$, we would like to estimate probability distribution $p(s_t|u_{1:t}, z_{1:t})$. However, $p(s_t|u_{1:t}, z_{1:t})$ can be an arbitrary distribution, thus we leverage Monte Carlo principle to address this issue.

The idea of Monte Carlo-based visual localization is to represent the probability distribution 
$P(s_t|u_{1:t}, z_{1:t})$ with a set of $N$ particles. Each particle maintains an estimate of 6 DoF pose at time t: 

\[\cS_t = \{s^{[1]}_t, s^{[2]}_t, ..., s^{[N]}_t \}\]

\noindent with a set of corresponding weights: 

\[\mathcal{W}_t = \{w^{[1]}_t, w^{[2]}_t, ..., w^{[N]}_t \} \]

The poses of particles and their corresponding weights are respectively updated according to the motion $u_t$ (Section \ref{sec:motion_model}) and noisy measurement $z_t$ (Section \ref{sec:noisy_measurement}) at time $t$. In the Section \ref{sec:motion_model}, we also justify the use of a simple motion model. Figure \ref{fig:pf_diagram} shows an overview of MCVL.

\subsection{Motion model} 
\label{sec:motion_model}
\begin{figure}[h]
	\centering
	\includegraphics[width=0.6\textheight]{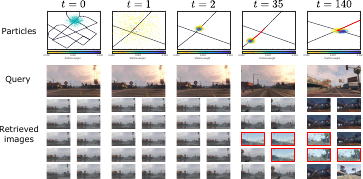}
	
	\caption{An example of MCVL. 1st and 2nd rows respectively are the particle distribution and query image, from 3rd to 7th rows are the retrieval results. Red bounding boxes indicate mistakes. The color bar describes the particle weights. Red lines are the camera pose predicted by MCVL. At the first frame, we randomly generates 1,000 particles with weight $0.001$, then particles which are inconsistent with measurements are vanished. Eventually, the ambiguity is resolved, and the particles track the movement of the vehicle. It is also worth mentioning that the query and retrieved images are from different time and weather conditions. In particular, the query images are from 6:26pm-cloudy, and the correct retrieved images are from 11:24am-rainy, 9:15pm-clear, and  1:28pm-sunny conditions.}
	\label{fig:pf_illustrative_example}
	
\end{figure}

In the autonomous driving scenario when navigating the roads of an urban area\footnote{In more ``localized" operations such as parking, where highly accurate 6 DoF estimation is required, it is probably better to rely on the INS.}, the motion of the car is fairly restricted, i.e., it largely stays on a road network on an approximately 2D plane\footnote{More fundamentally, the car is a non-holonomic system \cite{nonholonomicWiki}.} \cite{rubino2018practical}. 
While the road surface still allows significant scope for movement (cf.~Fig.~\ref{fig:concept}), relative to the size of the map, the motion of the car is rather limited\footnote{On uneven or hilly roads, accelerometers can be used to estimate the vertical motion, hence, VL can focus on map-scale navigation.}. This is echoed by the recent work of~\cite{sattler2018benchmarking}, who observed that there is ``\emph{lower variation in viewpoints as the car follows the same road}". Hence, a Monte Carlo scheme with a simple motion model suffices to track the 6 DoF rigid motion of the vehicle.
In many cities, the road networks are complex Euclidean graphs. In fact, it is well known that using (visual) odometry alone, it is possible to accurately track a car on a 2D road map \cite{brubaker2013lost}\footnote{The method of \cite{brubaker2013lost} will give ambiguous results on non-informative trajectories, e.g., largely straight routes. Hence, VL is still crucial.}. More fundamentally, this suggests that temporal continuity in the testing sequence (which is fully exploited by our method) strongly benefits VL.


Mathematically, for each particle, we model the noisy motion consisting of linear velocity $v^{[i]}_t$ and angular velocity $\psi^{[i]}_t$ as the following:

\begin{equation}
	\begin{matrix}
		v^{[i]}_t \sim \mathcal{N}(\mu_v, \Sigma_v) \\
		\psi^{[i]}_t \sim \mathcal{N}(\mu_{\psi}, \Sigma_{\psi})
	\end{matrix}
	\label{eq:velocity_sampling}
\end{equation}

\noindent where, $\mu_v$ and $\mu_{\psi}$ respectively represent the linear and angular velocities. The accelerations are modeled by the noise covariance matrices $\Sigma_v$ and $\Sigma_{\psi}$. For each particle, their motion in each time step is given by:

\[u^{[i]}_t = [v^{[i]}_t, \, \psi^{[i]}_t]^T\]

In practice, the $\mu_v$, $\mu_{\psi}$, $\Sigma_v$, and $\Sigma_{\psi}$ can be either manually tuned, or estimated from training data \cite{ko2009gp}.

While 3D positions can be easily updated by using simple additions, we convert two Euler angles to the Direction Cosine Matrix (DCM)~\cite{junkins2009analytical}, multiply two matrices and convert the result back to the Euler representation, to stay on the 3D manifold of valid rotations.
Let $\varphi(.)$ be a function that maps an Euler representation to DCM and  $\varphi^{-1}(.)$ is its inverse mapping. Our motion model for each particle is then given by:


\begin{equation}
	s^{[i]}_t = \begin{bmatrix} r^{[i]}_{t-1} +  v^{[i]}_t  \\ \varphi^{-1}\begin{pmatrix}\varphi(\psi^{[i]}_t).\varphi(\Omega^{[i]}_{t-1})\end{pmatrix} \end{bmatrix}
	\label{eq:motion_update}
\end{equation}

The experimental results will show that our motion model can properly handle the temporal evolution of camera pose in an image sequence. In case there is mismatch between the actual motion and the one predicted by the motion model, such as during emergency maneuvers, the discrepancy would be reflected in the enlarged covariance estimate and resolved once motion returns to within normal bounds.

\subsection{Estimating noisy measurements} 
\label{sec:noisy_measurement}
The similarity between the query and database images is calculated using $L$-2 distance:
\begin{equation}
	D(Q_t, I_i) = \begin{Vmatrix} \tau(Q_t) - \tau(I_i)\end{Vmatrix}_2^2
\end{equation}
\noindent where, $\tau(.)$ is the observation encoder (see section \ref{sec:obs_enc}). Afterwards, top $R$ database images with smallest distances are retrieved. Next, mean-shift algorithm is applied on $R$ retrieved images over the translational part of their poses. 
We then select the largest cluster, and calculate the mean of translation and rotation \cite{markley2007averaging}, which is viewed as a noisy measurement $z_t$ from the image query $I_t$. 

\subsection{Updating particle weights and Resampling}
\label{sec:update_weights_resampling}

For every particle, its weight is computed as the following:

\begin{equation}
	w^{[i]}_t = p\begin{pmatrix}z_t|s^{[i]}_t\end{pmatrix} \propto e^{-\frac{1}{2}(z_t - s^{[i]}_t)^T\Sigma_o^{-1}(z_t - s^{[i]}_t)} 
	\label{eq:weight_update}
\end{equation}

\noindent where, $\Sigma_o$ is a covariance matrix which describes the noise of the measurement obtained by observation encoder. Then, all particle weights are normalized to ensure their summation equal to $1$:

\begin{equation}
	\forall i, w^{[i]}_t = \frac{w^{[i]}_t}{\sum_{j=1}^{n} w^{[j]}_t}
	\label{eq:weight_normalize}
\end{equation}

Finally, we resample particles based on their weights by stochastic universal sampling \cite{whitley1994genetic}. This resampling step prevents the degeneracy problem, which can occur in the long-term localization scenario. 
Fig. \ref{fig:pf_illustrative_example} shows the filtering performed by our proposed method. 
At the first iteration, hypotheses are randomly generated. Hypotheses with small weights vanish if they are inconsistent with the noisy measurement. 
Finally, the ambiguity is resolved, and the particles successfully track the vehicle. It is worth noting that in the example shown, the query and retrieved images are from different times and weather conditions.

\subsection{Overall algorithm}

Algorithm \ref{alg:mcvl_algo} summarizes the proposed Monte Carlo-based visual localization method. The critical benefit of maintaining a set of particles is to leverage Monte Carlo principle to approximate the probability distribution $P(s_t | u_{1:t}, z_{1:t})$. As the number of particles $N$ are sufficiently large, this set of particles are equivalent to a representation of probability density function. The motion model ensures the temporal smoothness of the vehicle's movement.

\begin{algorithm}
	\caption{Monte Carlo-based Visual Localization (MCVL)}
	\begin{algorithmic}[1]
		\Function{MCVL}{$\cS_{t-1}, \cW_{t-1}, Q_t$}
		\State $\overline{\cS_t} = \cS_t = \emptyset$
		\State $\overline{\cW_t} = \cW_t = \emptyset$
		\State Estimate $z_t$ from $Q_t$ (Section \ref{sec:noisy_measurement} )
		\For{$i = 1$ to $N$}
		\State sample $v^{[i]}_t$ and $\psi^{[i]}_t$ as (\ref{eq:velocity_sampling})
		\State compute $s^{[i]}_t$ as (\ref{eq:motion_update})
		
		\State compute $w^{[i]}_t$ as (\ref{eq:weight_update})
		\State $\overline{\cS_t} = \overline{\cS_t} + <s^{[i]}_t>$
		\State $\overline{\cW_t} = \overline{\cW_t} + <w^{[i]}_t>$
		\EndFor
		
		\State normalize $w^{[i]}_t$ as (\ref{eq:weight_normalize})
		\For{$i = 1$ to $n$}
		\State draw $i$ with probability $w^{[i]}_t$ (Section \ref{sec:update_weights_resampling})
		\State add $s^{[i]}_t$ to $\cS_t$
		\State add $w^{[i]}_t$ to $\cW_t$
		\EndFor
		
		\State \textbf{return} $\mathcal{S}_t$
		\EndFunction
	\end{algorithmic}
	\label{alg:mcvl_algo}
\end{algorithm}

\section{Collecting synthetic dataset using G2D \cite{Dzung18G2D}} \label{sec:g2d_and_dataset}
We construct a synthetic dataset from the computer game Grand Theft Auto V (GTA V). There are several softwares \cite{richter2017playing,krahenbuhl2018free,Dzung18G2D} which support researchers in collecting image data from GTA V. However, G2D \cite{Dzung18G2D} is selected, because it is able to produce an accurate ground truth 6 DoF camera pose for every image as well as provide functions to manipulate environmental conditions in the game. Note that there is an increasing recognition of the value of synthetic data towards building autonomous driving systems \cite{tremblay2018training}.

A brief overview of G2D is described in Section \ref{sec:g2d}, and further technical details can be found in \cite{Dzung18G2D}. Section \ref{sec:synthetic_dataset} is about our synthetic dataset for visual localization. More information regarding G2D and synthic dataset can be found in the project page: \url{https://sites.google.com/view/g2d-software/home}.

\subsection{G2D: From GTA to Data} \label{sec:g2d}

G2D aims to record two things from GTA V: images and their corresponding camera poses. 
To make the task of image collection easier and as automatic as possible,
two types of camera trajectories are used in G2D: \textit{sparse} and \textit{dense} trajectories. 
The sparse trajectory consists of a set of user-define vertices (positions on the ``top down" 2D map of the virtual environment), along with the order of visitation. 
G2D provides a function for users to manually select the coordinate locations of vertices on the 2D in-game map, and then users define the order of vertex visitation. 
Based on those information, a sparse trajectory is created by G2D. 
The protagonist in the game automatically traverses the user-defined sparse trajectory, during which the 6 DoF camera pose is recorded at \textbf{60 frames per second}\footnote{Based on Intel i7-6700 @ 3.40GHz, RAM 16GB, NVIDIA GeForce GTX 1080 Ti and the highest graphical configuration for GTA V.}, leading to a dense trajectory. This reduces user effort and enhances repeatability.
Finally images are collected by retracing a continuous path along specified by the dense trajectory. Each image captures the scene as observed from the gameplay camera.
This step is also performed automatically. 
Since the 6 DoF pose of the gameplay camera is recorded at 60 frames per second, the collected dataset is guaranteed in the standard video rate.

G2D also allows the user to manipulate in-game conditions such as weather, time of day, and traffic density. This manipulation is done before the image collection phase, which allows the user to capture images under different conditions from the exact same set of 6 DoF camera poses as specified by the dense trajectory.

\subsection{Synthetic data} \label{sec:synthetic_dataset}
\begin{figure*}[h]
	\centering
	\includegraphics[width=0.8\textwidth]{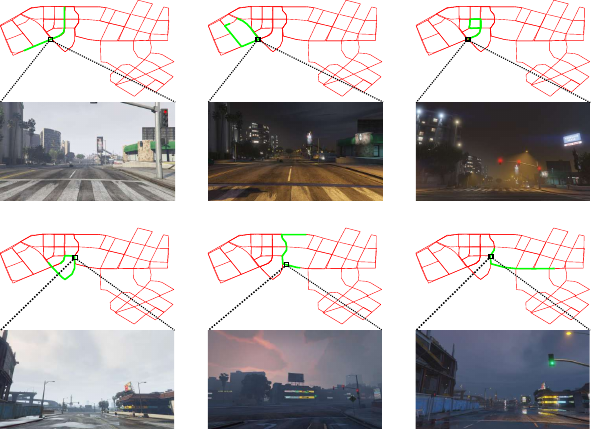}
	
	\caption{Our synthetic dataset simulates image sequences collected from multiple vehicles under different environmental conditions. The red lines indicate the road network of a part of a city, while the green lines indicate a few of the trajectories taken by different data collection vehicles. Sample frames at similar locations are shown---note the differences in \emph{pose} and environmental conditions between the images. Testing data will also be an image sequence recorded from a vehicle along the road network.}
	\label{fig:concept}
\end{figure*}

\begin{figure*}[h]
	\centering
	\includegraphics[width=0.8\textwidth]{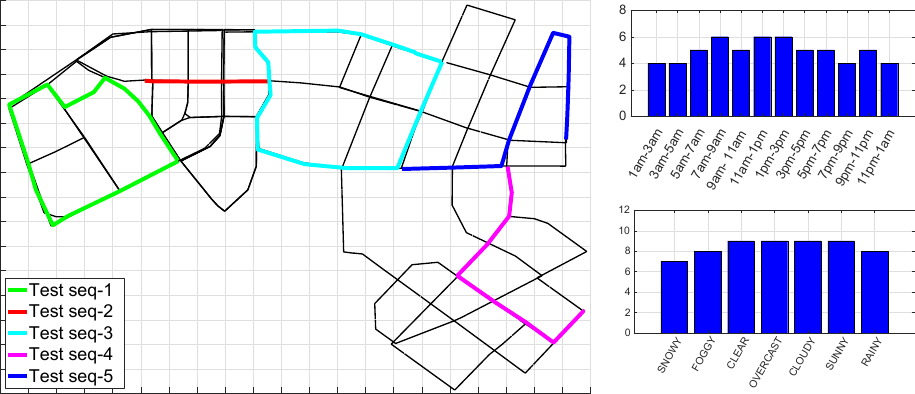}
	\caption{Bird's eye view and distribution of environmental conditions of our synthetic dataset. We simulate that there are 59 vehicles running in different routes, distinct time and weathers. Grid lines are every 100m. In training set, the coverage area is about 3.36km$^2$, the time and weather conditions are uniformly distributed. The statistics of testing sequences is shown in Table \ref{tab:stat_synthetic_dataset}}
	\label{fig:synthetic_dataset_birdeyeview}	
\end{figure*}

\begin{figure*}[h]
	\centering
	\includegraphics[width=0.8\textwidth]{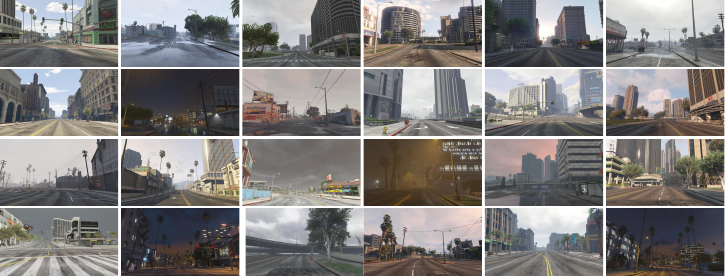}
	\caption{Training (1st-3rd rows) and testing samples (4th row) in our synthetic dataset.}
	\label{fig:synthetic_data_sample}
\end{figure*}

We simulate that there are 59 vehicles which run independently in different routes and environmental conditions. Figure \ref{fig:concept} shows the setting of our synthetic dataset. The bird's-eye view of our synthetic dataset is illustrated in Figure \ref{fig:synthetic_dataset_birdeyeview}, the coverage area is about 3.36km$^2$. Also, those vehicles are simulated to run in different times of day and weather conditions. The distributions of times and weather conditions are shown in two histograms of Figure \ref{fig:synthetic_dataset_birdeyeview}. The times and weathers in training sequences are uniformly distributed from 1am to 11pm, and in 7 different weather conditions (snowy, foggy, clear, overcast, cloudy, sunny and rainy). Five sequences in different times and weathers are also collected for the testing phase. The statistics information and trajectory of the testing sequences are described in Table \ref{tab:stat_synthetic_dataset} and Figure \ref{fig:synthetic_dataset_birdeyeview} respectively. We sub-sampled on the training sequences at 2Hz and the testing sequences at 4Hz. Examples from training and testing sequences are shown in Figure \ref{fig:synthetic_data_sample}.

\begin{table}
	\centering
	\begin{tabular}{l|l|l|l}
		\hline
		Sequences & \# images & Time \&  & Traversal \\
		&            & Weather & distance \\
		\hline
		\hline
		
		Test seq-1 & 1451 & 9:36 am, snowy & 1393.03m \\
		
		Test seq-2 & 360 & 10:41 pm, clear & 359.74m \\
		
		Test seq-3 & 1564 & 11:11 am, rainy & 1566.93m \\
		
		Test seq-4 & 805 & 6:26 pm, cloudy & 806.56m \\
		
		Test seq-5 & 1013 & 3:05 pm, overcast & 1014.91m \\
		
		\hline
	\end{tabular}
	\caption{Statistics of the testing sequences in the synthetic dataset}
	\label{tab:stat_synthetic_dataset}
\end{table}

\section{Experiments} \label{sec:experiments}

To validate the performance of our proposed VL methods, we conduct experiments on synthetic as well as real datasets.
For quantitative results, the translation error is calculated as the Euclidean distance $||c_{est} - c_{gt}||_2$. The orientation error $|\alpha|$ is computed as the angular difference $2 \, cos(|\alpha|) = trace(R_{gt}^{-1} \, R_{est})$ between estimated and ground truth camera rotation matrices $R_{est}$ and $R_{gt}$. 

A query frame is deemed as \textit{correctly localized} if its predicted pose is within a threshold of ($X m,Y^\circ$) from its ground truth pose. We define following thresholds: ($1m, 5^\circ$), ($5m, 10^\circ$), ($10m, 20^\circ$), ($15m, 30^\circ$),($20m, 40^\circ$), and ($50m, 100^\circ$)

\subsection{Implementation details} \label{sec:implementation_details}

In the observation encoder, we extract SIFT feature at 4 different scales with region width of 16, 24, 32 and 40, over a grid with spacing of 2 pixels. 
The SIFT features are embedded and aggregated using VLAD to obtain a single vector of length $16,384$, which is then projected to a $4,096$ dimensional space via PCA, whitened and L2-normalized.
To train the visual vocabulary, we randomly sample $5,000,000$ SIFT features in sequences ``28/11/2014, 12:07:13" and ``02/12/2014, 15:30:08" of Oxford RobotCar \cite{OxfordRobotCar}, and train a visual vocabulary size of $K = 128$.

In the visual localization with HMM, we set $\bv_{\text{max}} = 5$ and $\sigma = 0.06$. The number of place hypotheses are $|\cH_t| = 20$. the initial belief $\bp_0$ is uniformly initialized. This approach is termed as HMM in following sections.

In MCVL, in case prior information of initial camera pose is unknown, ones typically initialize particles according to uniform distribution. However, since a rough camera pose can be inferred from image retrieval, it is reasonable to assume that high (and small) probability for initial camera pose closer (and further) to that rough camera pose. Therefore, particles are initialized from Gaussian distribution with the mean from the noisy measurement in the first frame.
The covariance matrices for initializing 3D location $r^{[i]}_{t=1}$ and orientation $\Omega^{[i]}_{t=1}$ respectively are $diag([10, 10, 10]^T)$ and $diag([0.001, 0.001, 1]^T)$. The parameters of motion model are set as the following: $\Sigma_o= diag([5, 5, 5, 0.0001, 0.0001, 0.001]^T)$, $\Sigma_v = diag([1,1,0.01]^T)$, $\Sigma_{\psi} = diag([0.0001 ; 0.00001 ; 0.01]^T)$, $\mu_v = [0.1, 0.1, 0.01]^T$, $\mu_{\psi} = [0.001, 0.00001, 0.01]^T$, where $diag$ is a function that outputs a square diagonal matrix with the elements of input vector on the main diagonal. The number of particles are fixed to $1000$. The numbers of retrieved images $R=20$. 

We select CNN-based 6-DoF camera pose regression approaches as our competitors: PoseNet \cite{kendall2015posenet}, MapNet \cite{brahmbhatt2018mapnet} and MapNet+PGO \cite{brahmbhatt2018mapnet}. Following suggestion of \cite{brahmbhatt2018mapnet}, PoseNet uses ResNet-34 \cite{he2016resnet}, adds a global average pooling layer, and parameterizes camera orientation as logarithm of a unit quaternion. MapNet+PGO employs pose graph optimization (PGO) to fuse the absolute poses (produced by MapNet) and relative odometry (from visual odometry) during the inference. PoseNet is implemented in Tensorflow, MapNet and MapNet+PGO's implementations are provided by the authors. Parameters of PoseNet and MapNet are set as the suggestion of the authors.

\subsection{Datasets}

We conduct the experiment using our synthetic dataset (see Section \ref{sec:synthetic_dataset}), and Oxford RobotCar \cite{OxfordRobotCar}. For Oxford RobotCar, we follow the configuration suggested by MapNet \cite{brahmbhatt2018mapnet}. The split of training and testing sequences are summarized in Table \ref{tab:stat_robotcar}. The training and testing sequences are recorded in different dates, which ensure a significant difference in their appearances. The experiment is conducted on the alternate and full routes with the length of 1 km and 10 km respectively. MapNet \cite{brahmbhatt2018mapnet} is initially trained with the training set, and then fine-tuned in an unlabeled dataset, PoseNet \cite{kendall2015posenet} is trained with the training set, and our methods only use training set as the database $\cD$.

\begin{table}

	\centering
	
	\begin{tabular}{|c|l|l|}
		\hline
		Route & Purpose & Recorded  \\
		\hline
		\hline
		
		\multirow{5}{4em}{Alternate route (1 km)} & Training & 26/6/2014, 9:24:58 \\
		& Training & 26/6/2014, 8:53:56  \\
		& Unlabeled & 14/5/2014, 13:50:20  \\
		& Unlabeled & 14/5/2014, 13:46:12  \\
		
		& Query & 23/6/2014, 15:41:25 \\
		\hline
		\hline
		\multirow{4}{4em}{Full route (10 km)} & Training & 28/11/2014, 12:07:13 \\
		& Training & 02/12/2014, 15:30:08 \\
		& Unlabeled & 12/12/2014, 10:45:15  \\
		
		& Query & 09/12/2014, 13:21:02  \\
		\hline
	\end{tabular}

	\caption{The split of training and testing sequences in Oxford RobotCar dataset.}
	\label{tab:stat_robotcar}
	\vspace{-2em}
\end{table}

\subsection{Comparison between VLAD and Bag of Word (BoW)}
\begin{figure*}[h]
	\centering
	\mbox
	{		
		\subfloat[][]{
			\includegraphics[width=0.48\textwidth]{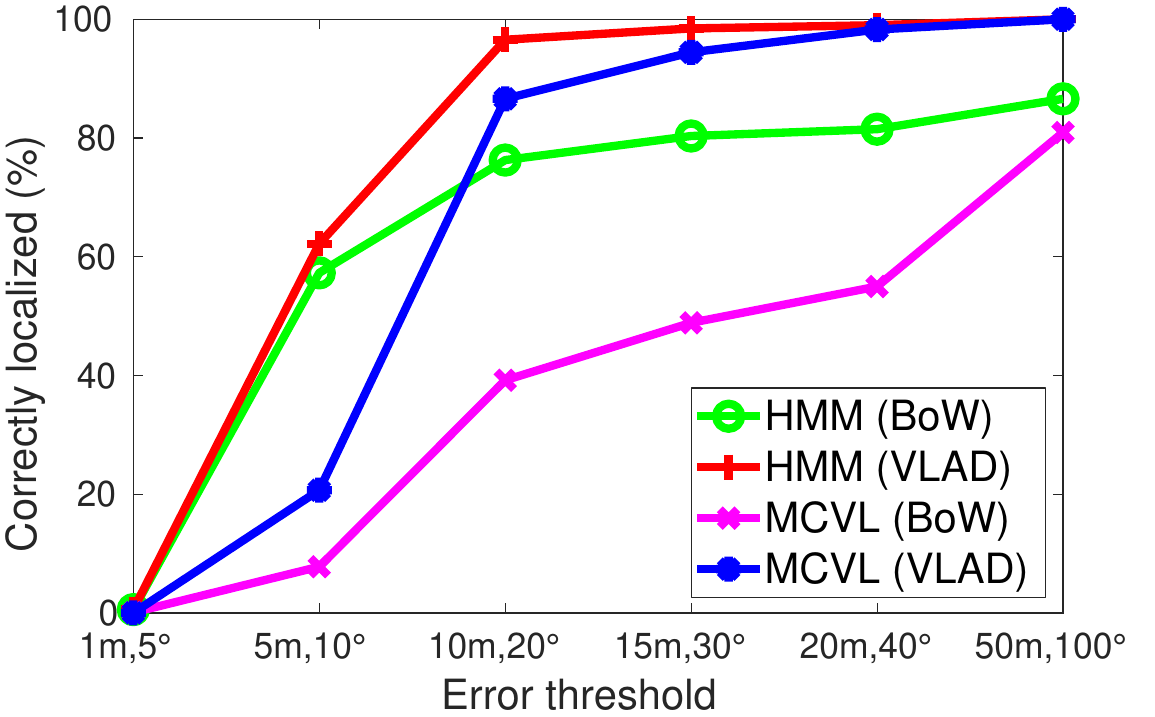}
			\label{fig:robotcar_recall:vlad_bow}}
		
		\subfloat[][]{
			\includegraphics[width=0.48\textwidth]{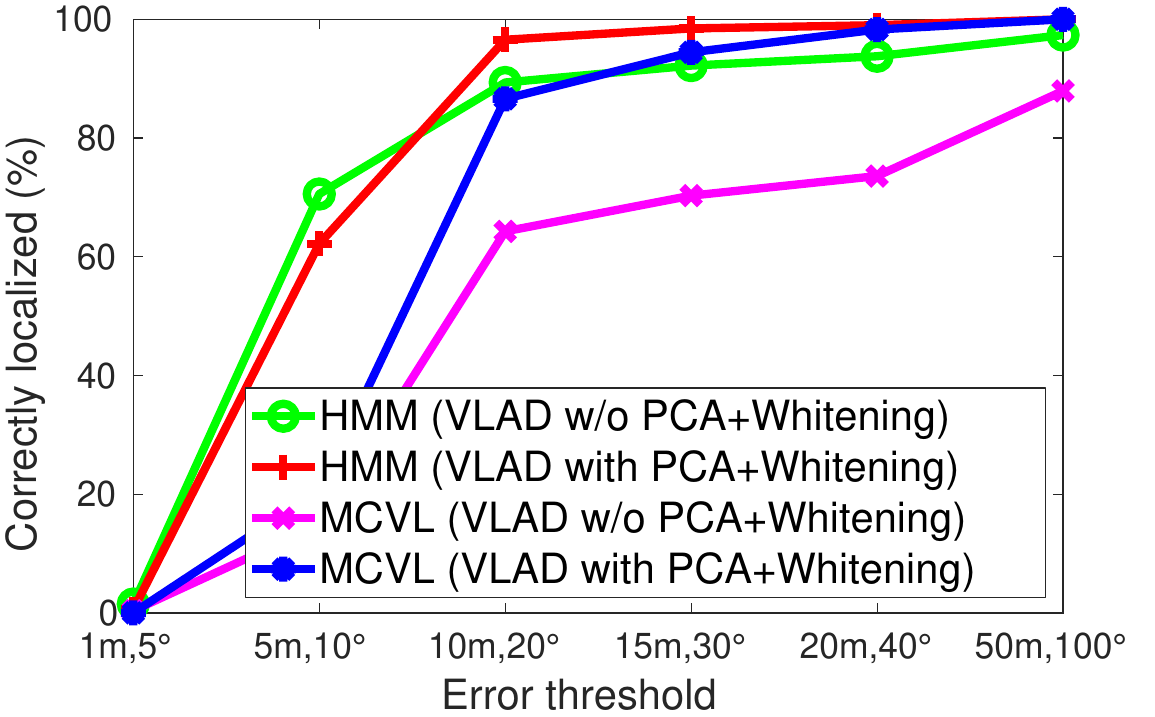}
			\label{fig:robotcar_recall:vlad_pca}}
		
	}

	\caption{Percentage of correctly localized query frames in Oxford Robotcar. \textbf{(a)} Comparison between VLAD and BoW, \textbf{(b)} The benefit of post-processing}
	\label{fig:robotcar_vlad_bow_pca}
	\vspace{-2em}
\end{figure*}

Figure \ref{fig:robotcar_recall:vlad_bow} compares the localization errors between VLAD and BoW. For BoW, the visual vocabulary of size $4,096$ is trained using the $5,000,000$ SIFT features as described in Section \ref{sec:implementation_details}. For each image, we also extract SIFT features in the same setting as Section \ref{sec:implementation_details}, the RootSIFT normalization (Section \ref{sec:obs_enc}) is subsequently performed. BoW representation is formed for each image, it is then followed by the $L$2 normalization: $x_i = x_i / ||x_i||_2$. Here,  since VLAD is a higher order representation than BoW, it significantly outperforms BoW.

\subsection{The benefit of post-processing}
Section \ref{sec:obs_enc} claims using PCA and whitening (post-processing) can reduce the impacts of background features. Figure \ref{fig:robotcar_recall:vlad_pca} verifies that this post-processing indeed  improves the localization accuracy. Note that without PCA projection, the dimensionality of each image representation is $16,384$ which also slows down the inference time.

\subsection{Sum and democratic aggregation}
\begin{figure*}[h]
	
	\centering
	
	\includegraphics[width=0.96\textwidth]{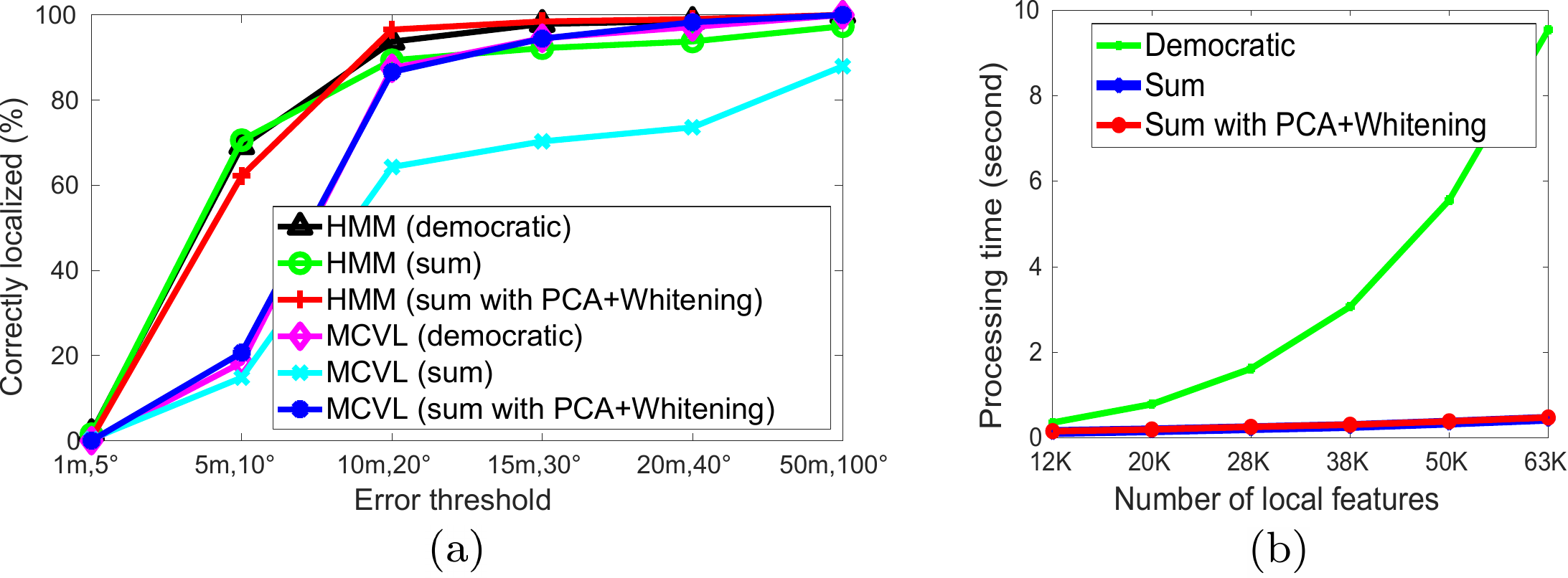}
	\label{fig:roborcar_sum_democratic:accuracy}
	
	\caption{Comparison between sum and democratic aggregations in Oxford RobotCar. \textbf{(a)} Localization accuracy, \textbf{(b)} Time of encoding the observation with respect to number of local SIFT features.}
	\label{fig:roborcar_sum_democratic}
\end{figure*}

This section verifies the practical value of using sum aggregation over democratic aggregation. For democratic aggregation, we use the implementation provided by \cite{jegou2014triangulation}. The purpose of democratic aggregation is to equalize the contribution of embedded vectors, reducing the impact of background features, since background features (e.g., tree, sky, road, etc) usually dominate the image. Therefore, from Figure \ref{fig:roborcar_sum_democratic}a, democratic aggregation outperforms sum aggregation. However, thanks to post-processing (PCA and whitening), it improves sum aggregation to be comparable to democratic aggregation. 

In addition, Figure \ref{fig:roborcar_sum_democratic}b shows that the processing time of democratic aggregation significantly becomes slower when the number of local SIFT features increase, while the processing time of sum aggregation combined with post processing remains also constant. This is because when democratic aggregation is performed, it needs to solve an optimization problem. By contrast, the post processing is only the matrix multiplication which can be processed efficiently in parallel. Hence, due to the computational complexity of democratic aggregation, its practicality is limited.
\subsection{Comparison to competitors}

\begin{figure*}[h]
	\centering
	\mbox
	{		
		\subfloat[][MapNet]{
			\includegraphics[width=0.24\textwidth]{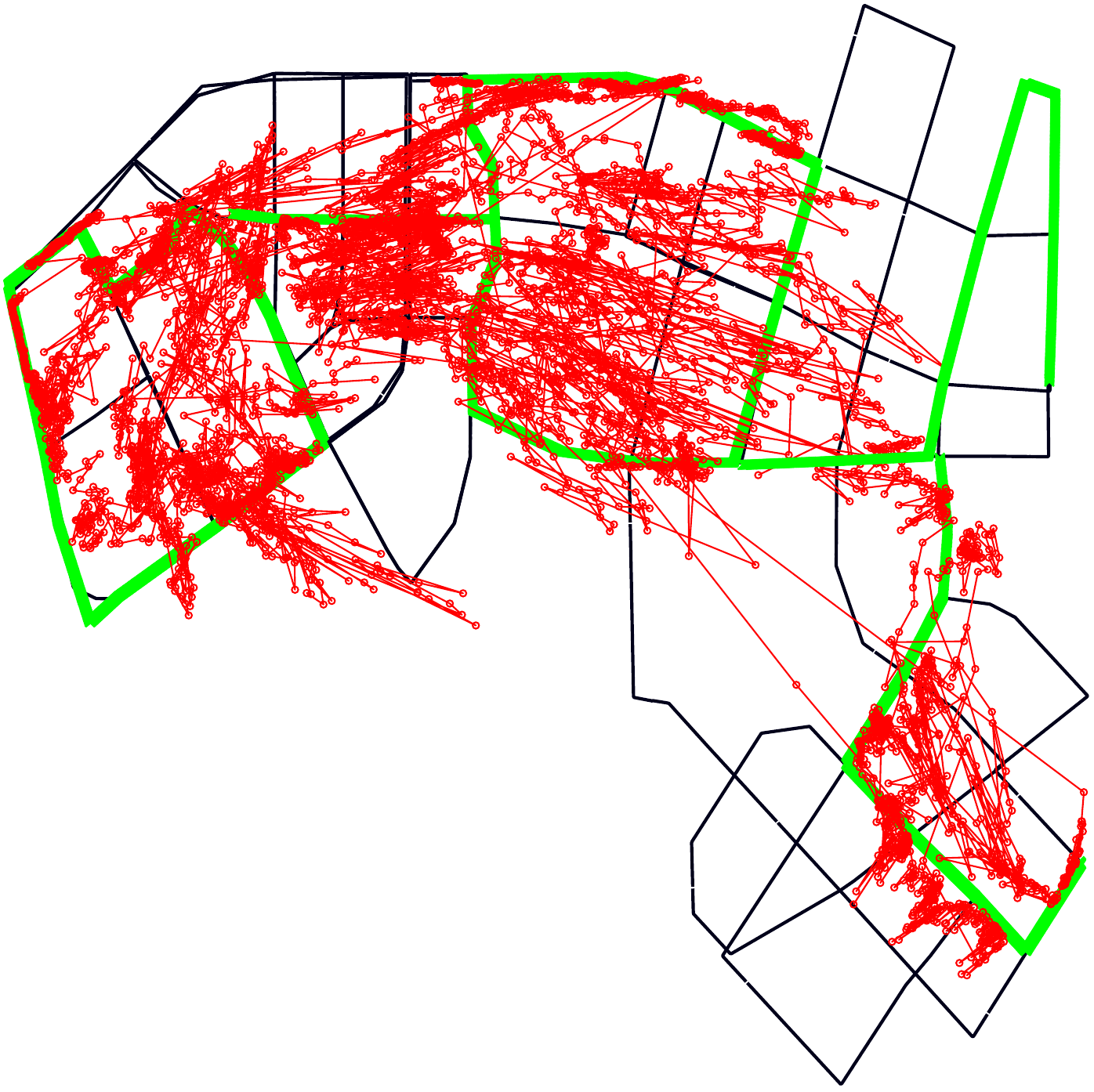}}
		\hspace{1cm}
		\subfloat[][HMM]{
			\includegraphics[width=0.24\textwidth]{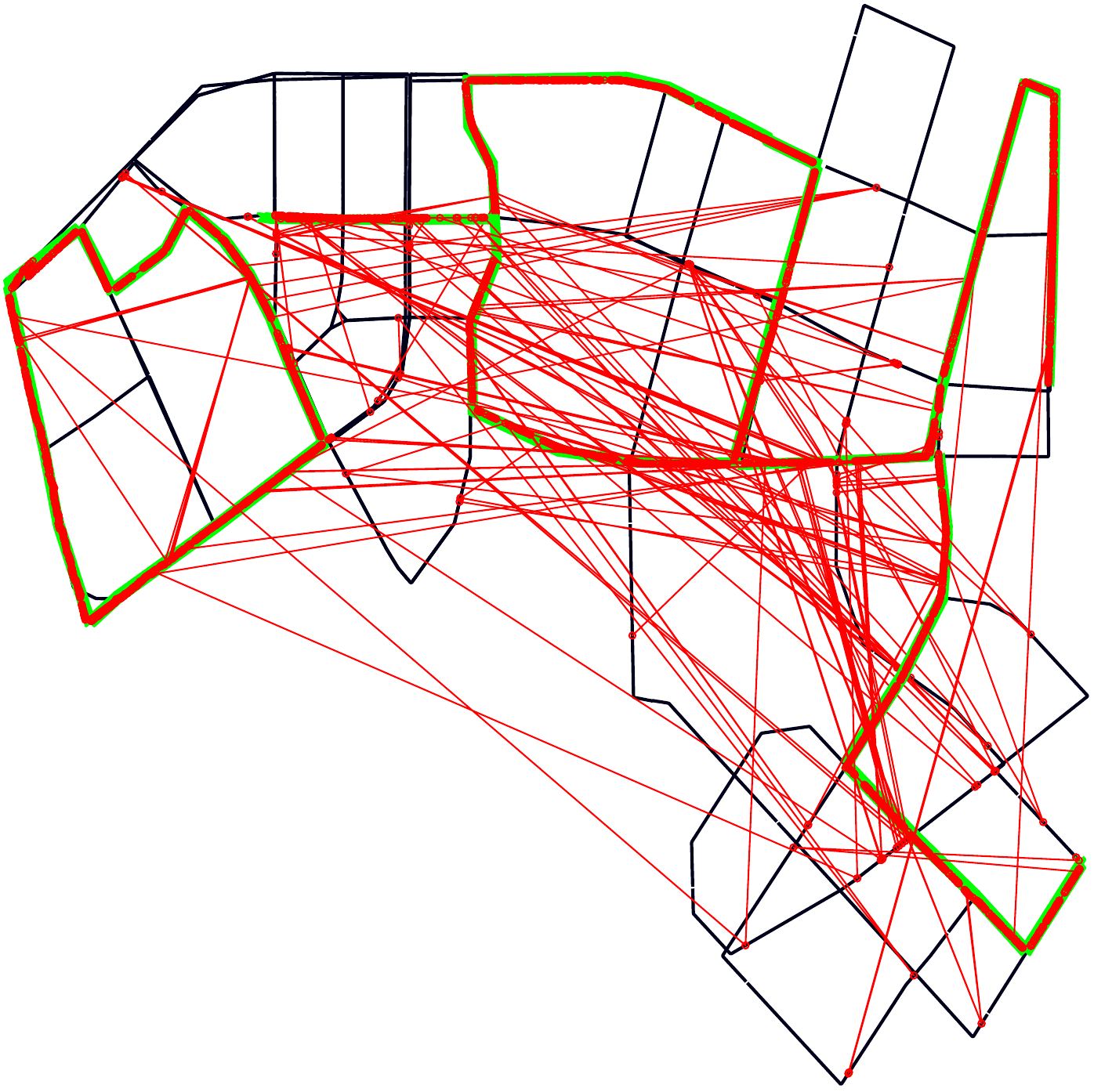}}
		\hspace{1cm}
		\subfloat[][MCL]{
			\includegraphics[width=0.24\textwidth]{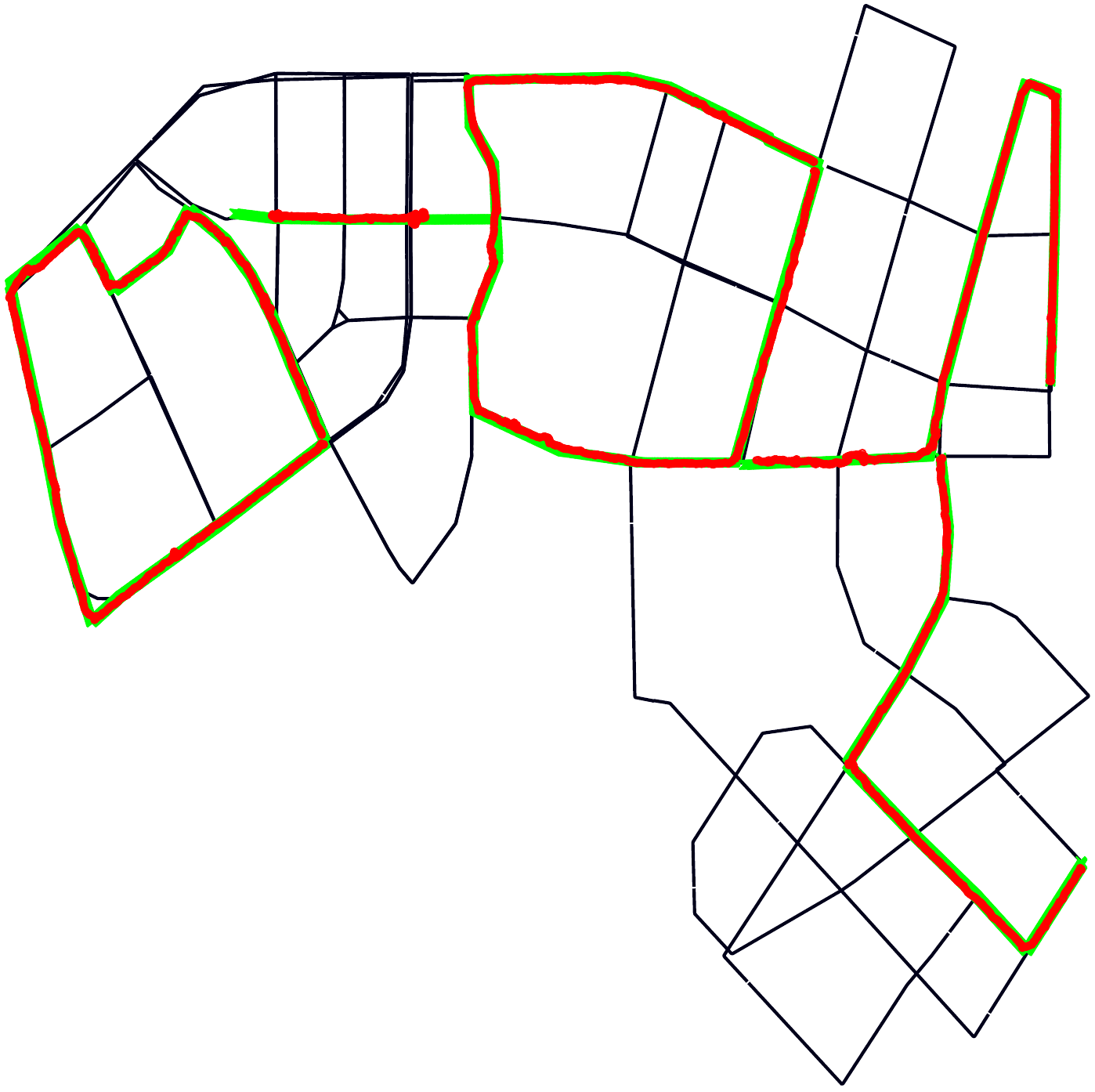}}
	}
	
	\caption{Results on our synthetic dataset. Green lines represented ground truth and predicted trajectories are given in red.}
	\label{fig:synthetic_result}
\end{figure*}

\begin{figure*}
	\centering
	\mbox
	{		
		\subfloat[][Test seq-01]{
			\includegraphics[width=0.32\textwidth]{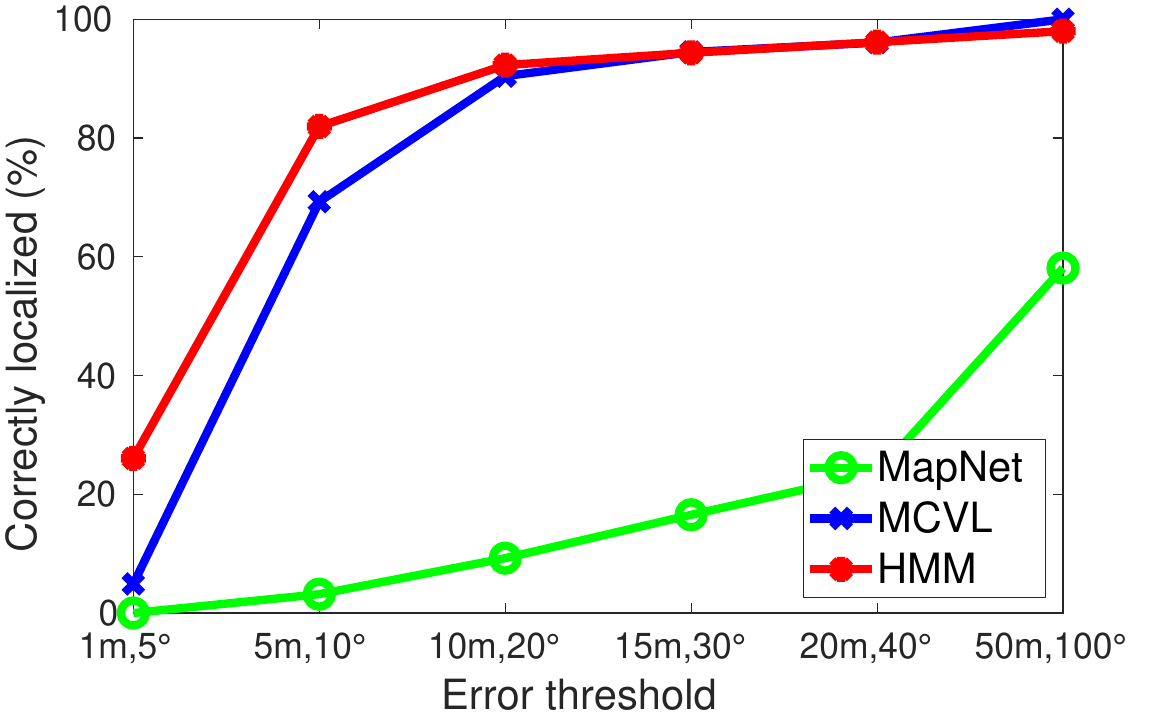}}
		
		\subfloat[][Test seq-02]{
			\includegraphics[width=0.32\textwidth]{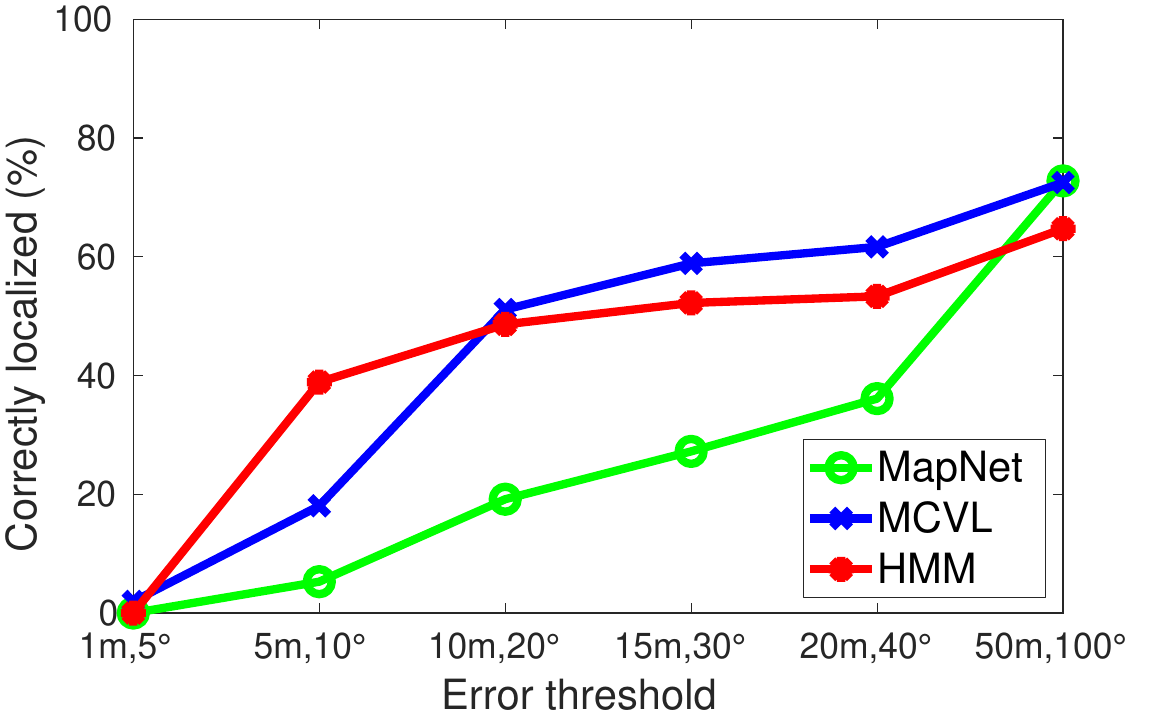}}
		
		\subfloat[][Test seq-03]{
			\includegraphics[width=0.32\textwidth]{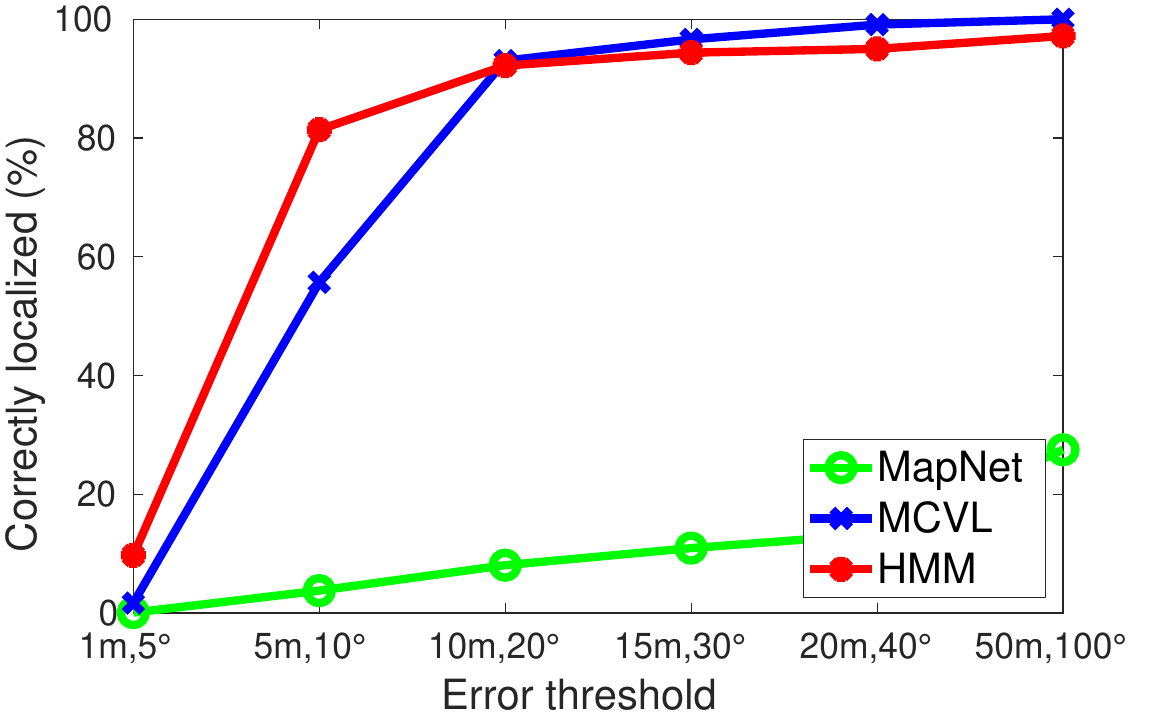}}
	}
	\mbox
	{		
		\subfloat[][Test seq-04]{
			\includegraphics[width=0.32\textwidth]{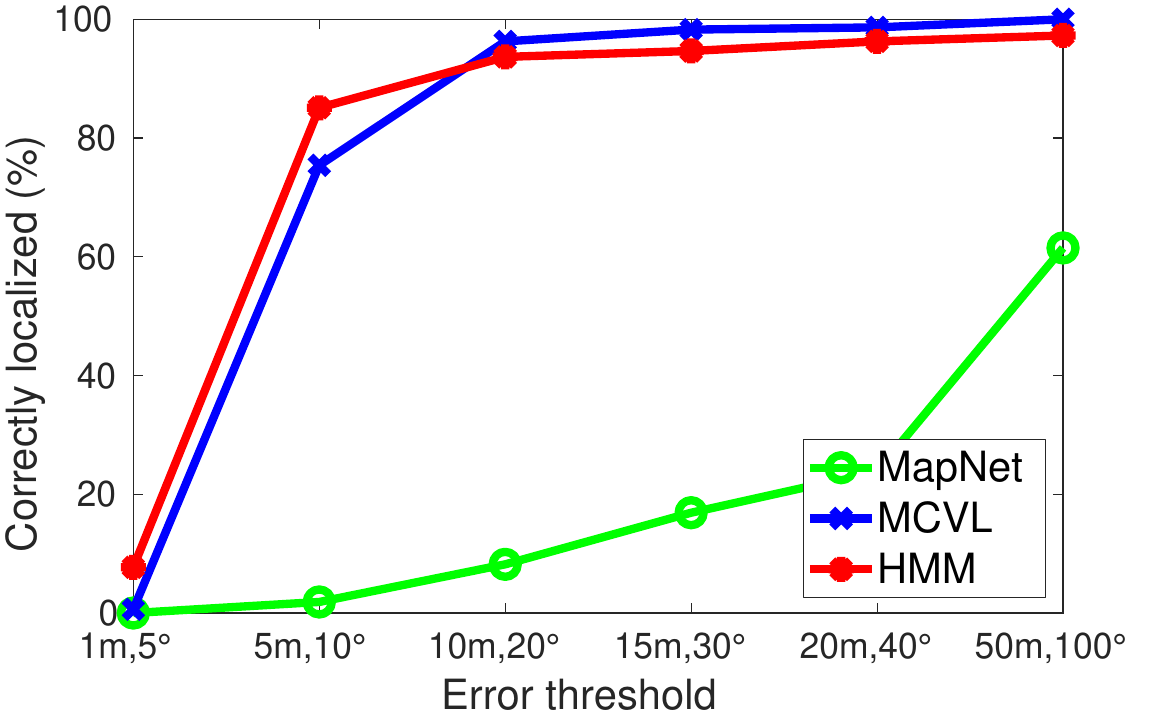}}
		
		\subfloat[][Test seq-05]{
			\includegraphics[width=0.32\textwidth]{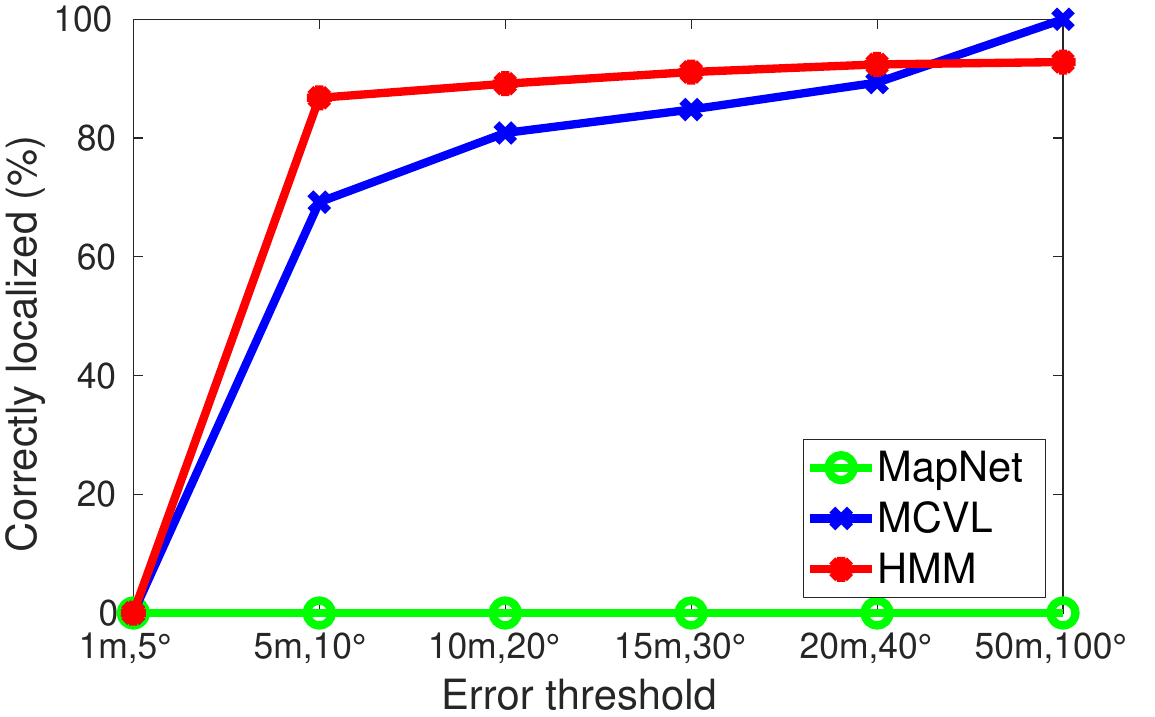}}
	}

	\caption{Percentage of correctly localized query frames in our synthetic dataset.}
	\label{fig:synthetic_recall}
	\vspace{-2em}
\end{figure*}

\paragraph{Synthetic dataset:} As can be seen from Figure \ref{fig:synthetic_result}, MapNet struggles to produce a reasonable result. 
This is because MapNet formulates the problem as an image to pose regression, whose underlying assumption of constant appearance is violated when the appearance of the environment changes. Moreover, repeated structures such as 
trees, sky, and road surfaces can leads to ambiguities for MapNet. 
In contrast, our observation encoder applies state-of-the-art normalization techniques to reduce the negative impact from repetitive objects. 
Hence as shown in Figure \ref{fig:synthetic_recall}, HMM and MCVL significantly outperform MapNet. Although HMM is slightly better than MCVL regarding the percentage of correctly localized queries, MCVL produces a more smooth trajectory than HMM (see Figure \ref{fig:synthetic_result}).

\paragraph{Oxford RobotCar:} \label{sec:experiment_robotcar}

\begin{figure*}[h]
	\centering
	\mbox
	{		
		\subfloat[][Alternate route (1km)]{
			\includegraphics[width=0.48\textwidth]{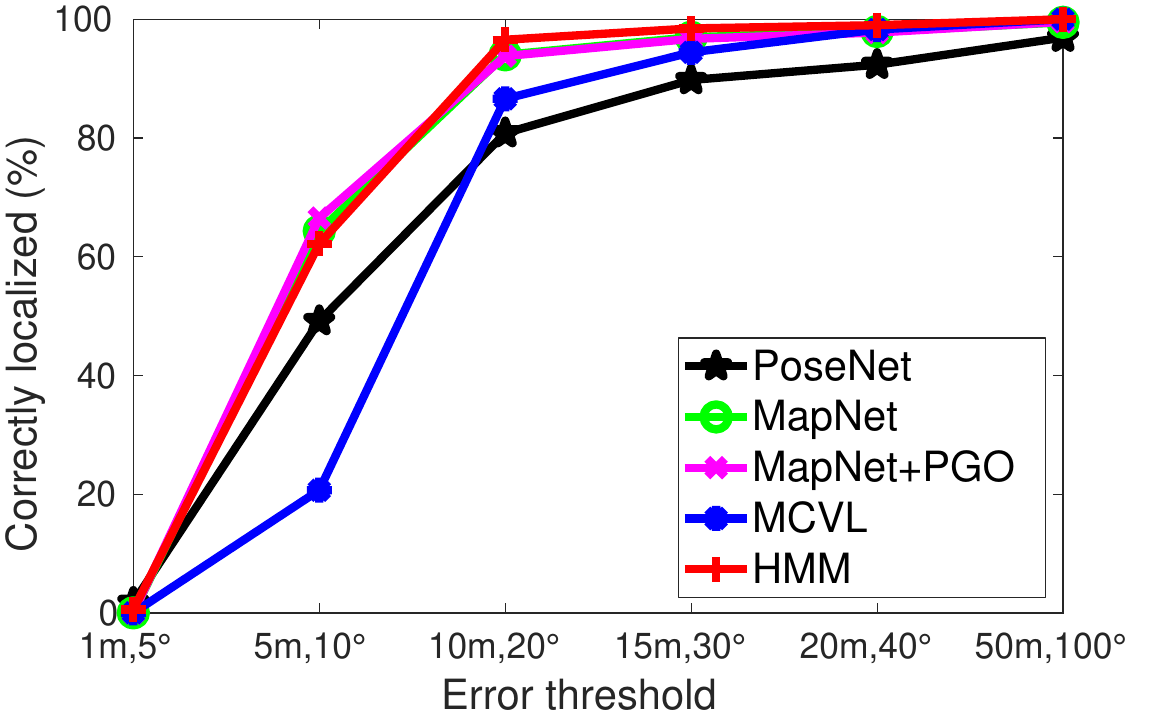}
			\label{fig:robotcar_recall:alternate}}
		
		\subfloat[][Full route (10km)]{
			\includegraphics[width=0.48\textwidth]{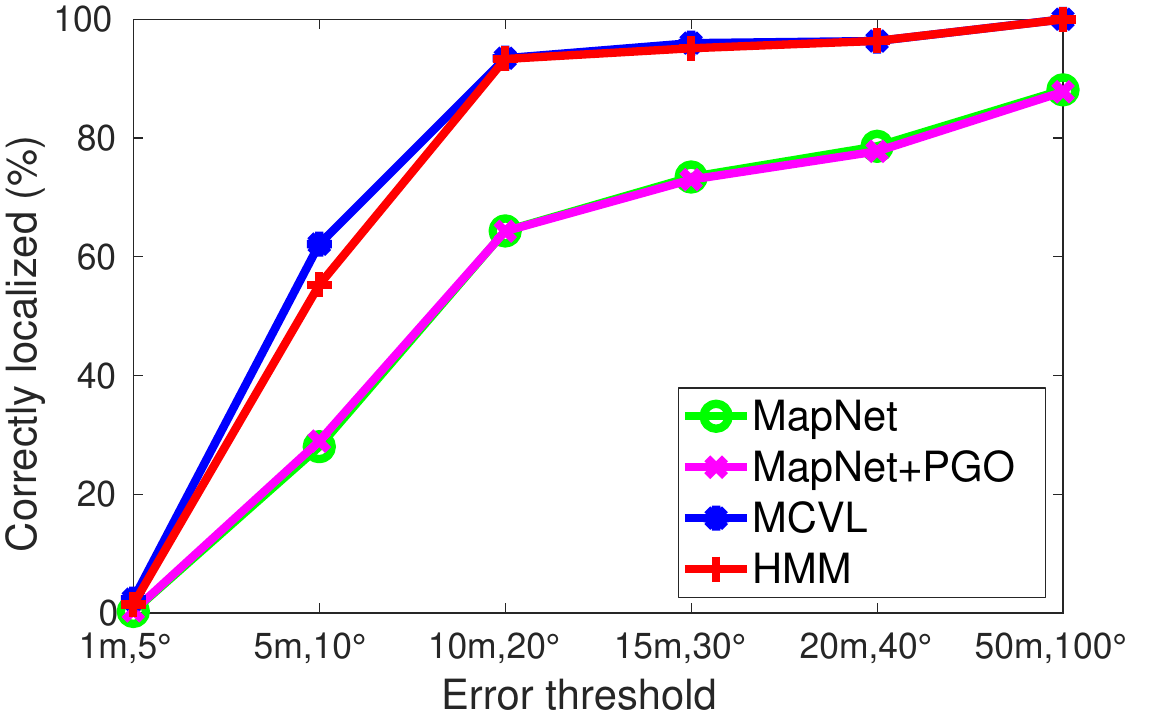}
			\label{fig:robotcar_recall:full}}
		
	}
	\caption{Percentage of correctly localized query frames in Oxford Robotcar.}
	\label{fig:robotcar_recall}
\end{figure*}

\begin{figure*}[h]
	\centering
	\mbox
	{		
		\subfloat[][PoseNet]{
			\includegraphics[width=0.22\textwidth]{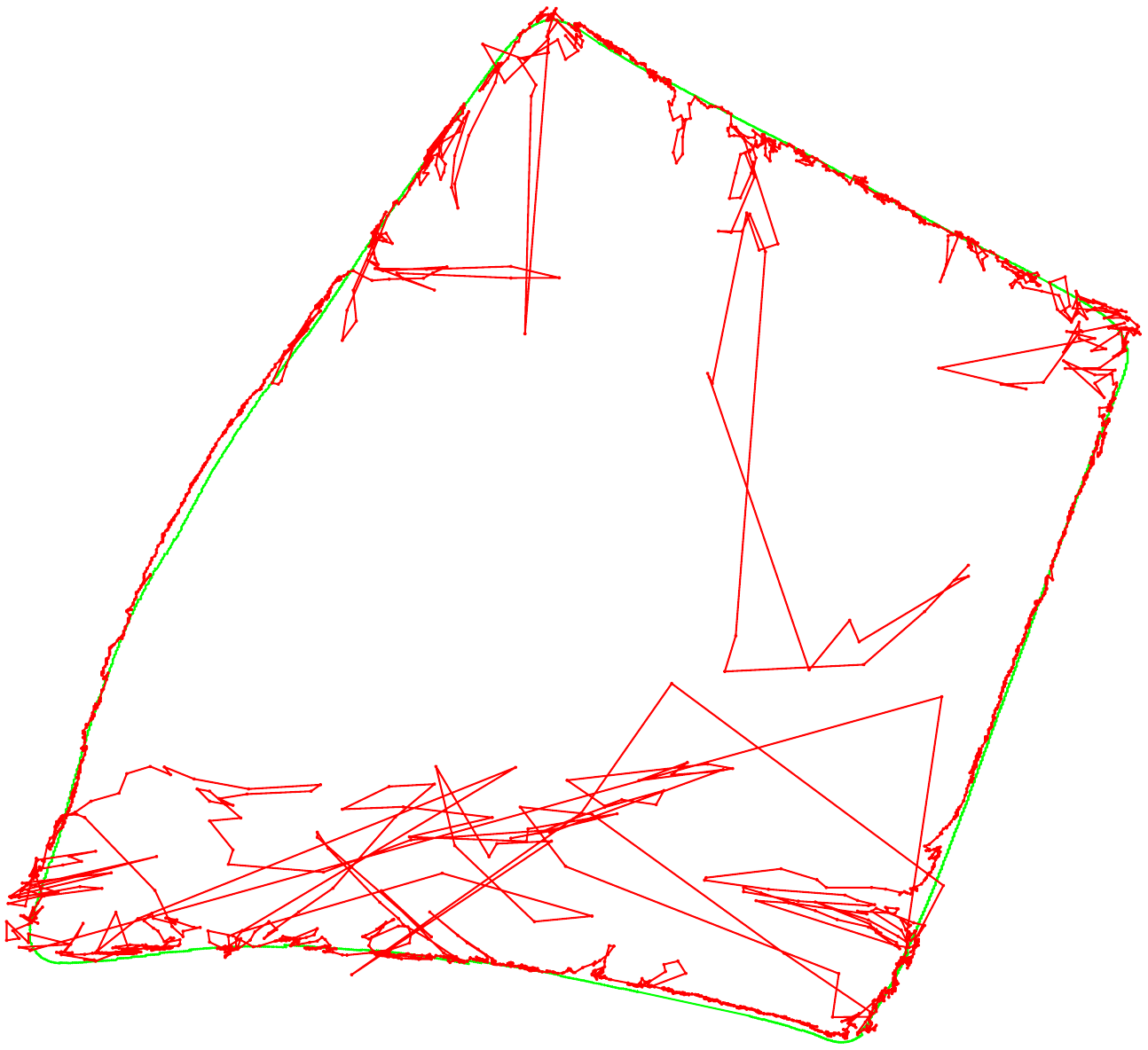}}
		\hspace{0.4cm}			
		\subfloat[][MapNet]{
			\includegraphics[width=0.22\textwidth]{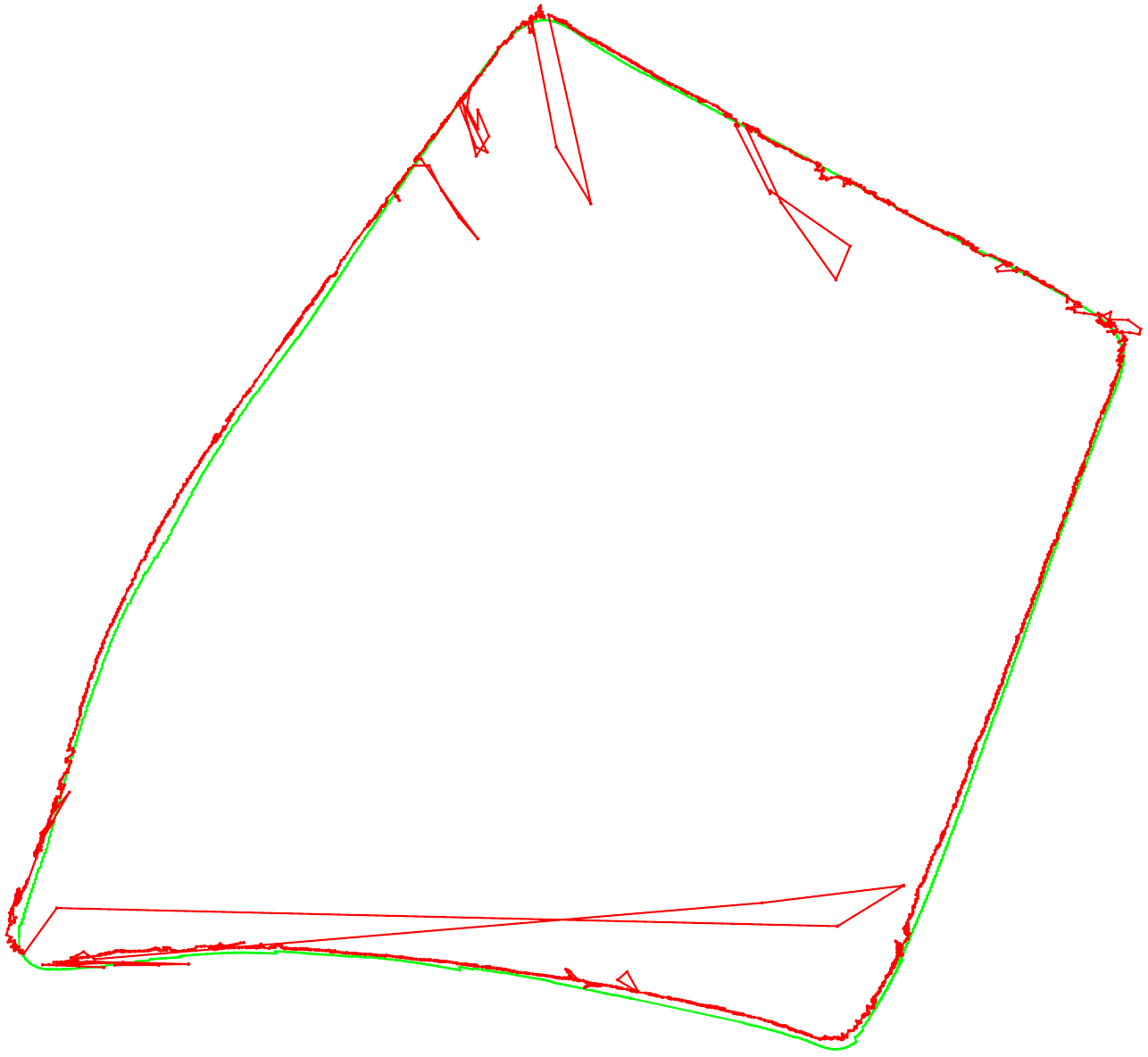}}
		\hspace{0.4cm}			
		\subfloat[][MapNet+PGO]{
			\includegraphics[width=0.22\textwidth]{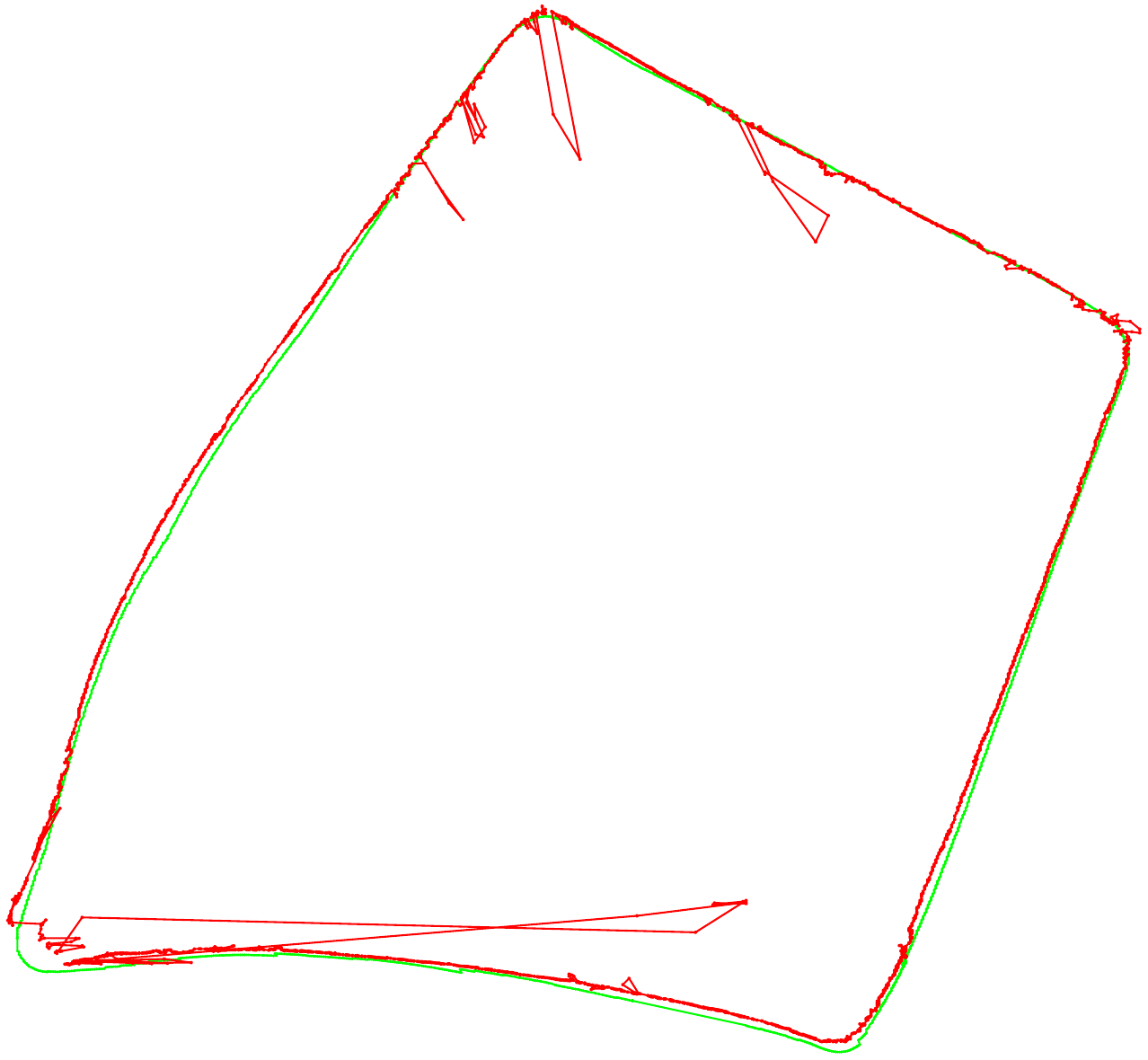}}
		
	}
	
	\mbox
	{
		\subfloat[][HMM]{
			\includegraphics[width=0.22\textwidth]{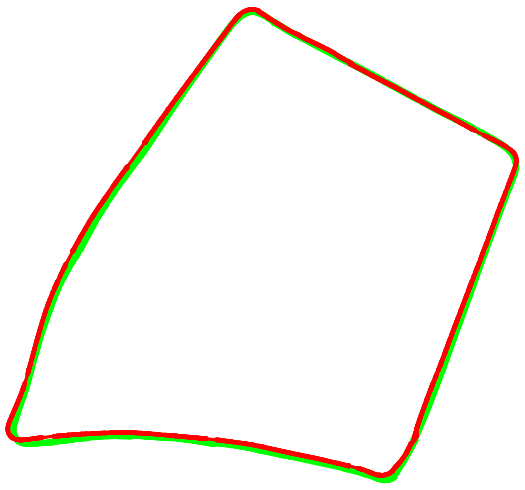}}
		\hspace{0.4cm}			
		\subfloat[][MCVL]{
			\includegraphics[width=0.22\textwidth]{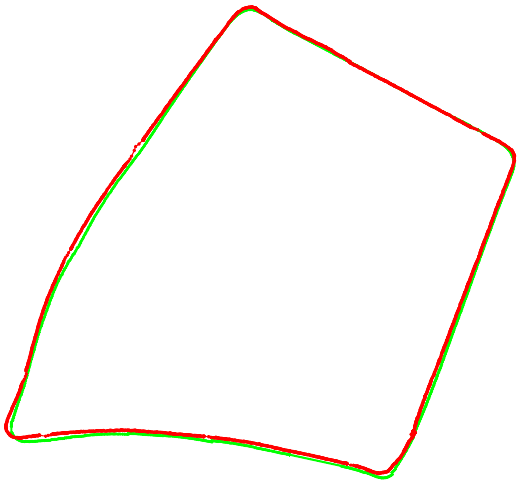}}
	}	
	\caption{Results on alternate route (1km) in Oxford RobotCar dataset. The green lines are ground truth, and red lines are predicted trajectories.}
	\label{fig:robotcar_result_alternate}
\end{figure*}

\begin{figure*}
	\centering
	
	\mbox
	{
		\subfloat[][MapNet]{
			\includegraphics[width=0.20\textwidth]{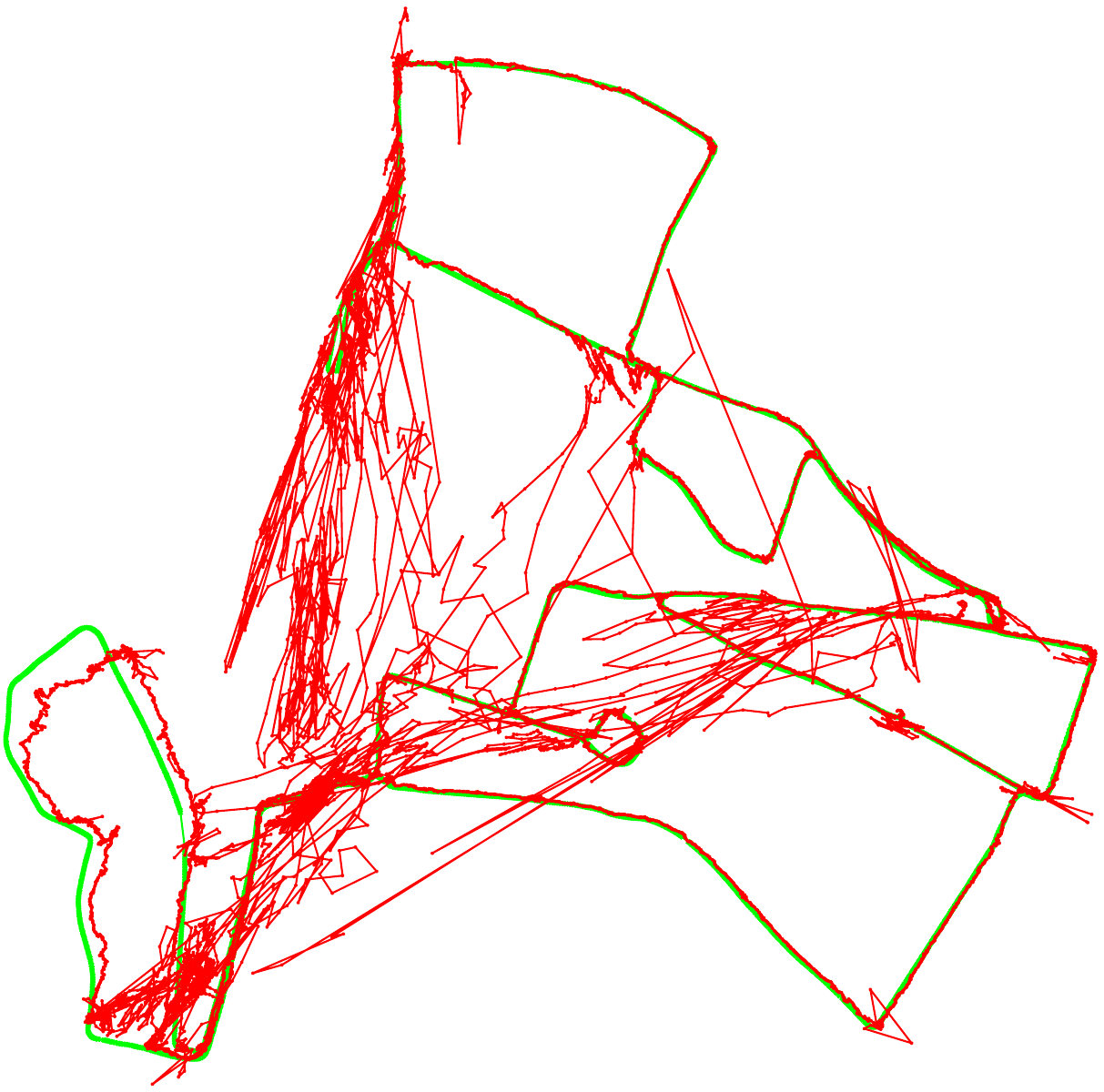}}
		
		\hspace{0.4cm}				
		\subfloat[][MapNet+PGO]{
			\includegraphics[width=0.20\textwidth]{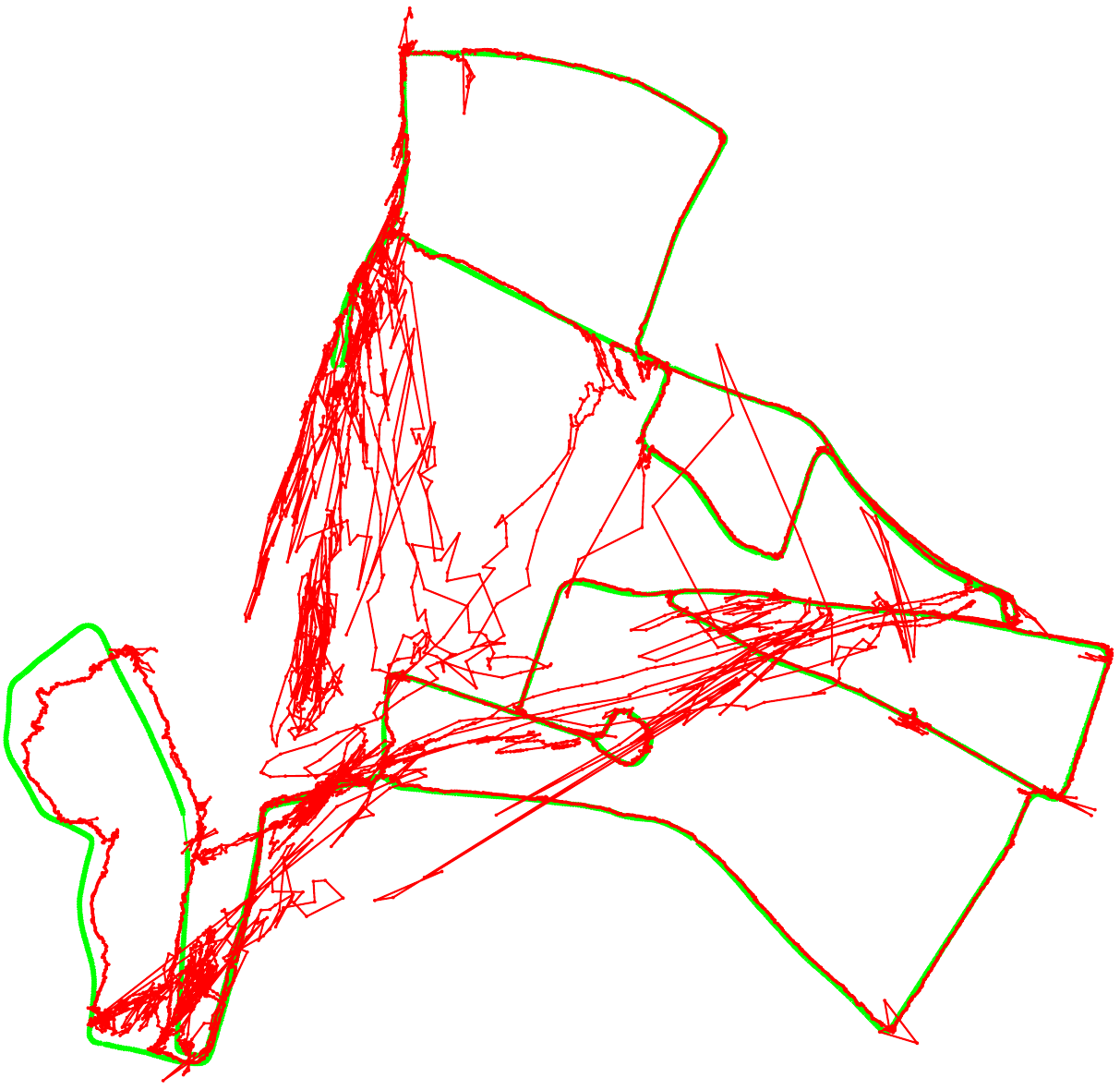}}	
		
		\hspace{0.4cm}	
		\subfloat[][HMM]{
			\includegraphics[width=0.20\textwidth]{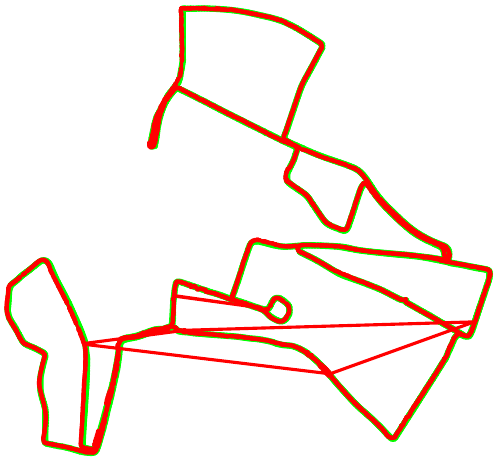}}	
		
		\hspace{0.4cm}	
		\subfloat[][MCVL]{
			\includegraphics[width=0.20\textwidth]{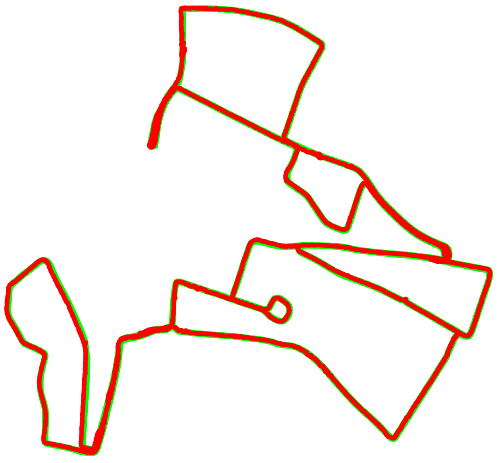}}	
	}	
	\caption{Results on full route (10km) in Oxford RobotCar dataset. The green lines are ground truth, and red lines are predicted trajectories.}
	\label{fig:robotcar_result_full}
\end{figure*}

We compare our proposed method to state-of-the-art approaches, i.e., PoseNet \cite{kendall2015posenet}, MapNet and MapNet+PGO \cite{brahmbhatt2018mapnet}. In particular, PoseNet directly regresses 6 DoF camera pose from an input image. MapNet receives videos as training data, hence its loss function minimizes absolute pose per image as well as the relative pose between consecutive images, it is then followed by a fine-tuning step on unlabeled data with their visual odometry (VO). MapNet+PGO, in the inference step, fuses the prediction of MapNet with VO by using pose graph optimization (PGO) to ensure the temporal smoothness.

Results on alternate route (1km) are shown in Figure \ref{fig:robotcar_recall:alternate}. MapNet, MapNet+PGO and MCVL share a comparable performance, which outperforms PoseNet and HMM. However, Figure \ref{fig:robotcar_result_alternate} shows that MapNet, MapNet+PGO and PoseNet are unable to output a smooth trajectory, while our methods (i.e., MCVL and HMM) can produce a smooth prediction. One possible reason which PGO is unable to produce a smooth prediction is there are some existing outliers from the predictions of both MapNet and VO which will probably make PGO stuck in a local minimum.

In the large-scale setting, which is full route (10km), PoseNet is not reported since it can not give a reasonable result. In general, we observe the results on full route are consistent with those on alternate route, i.e., HMM and MCVL output more smooth predictions, compared to MapNet and MapNet+PGO. Therefore, HMM and MCVL significantly outperform MapNet and MapNet+PGO in terms of percentage of correctly localized queries (see Figure \ref{fig:robotcar_recall:full}).

\section{Discussion \& Conclusion} \label{sec:discussion}
This paper shows that using hand-crafted features and embedding technique combined with a temporal probabilistic framework (i.e., Hidden Markov Model or Monte Carlo Localization) can provide a reasonable visual localization accuracy, compared to deep learning pose regression approaches. In addition, the experiments suggest future directions for visual localization: i) instead of investigating deep learning-based pose regression methods, exploiting deep learning-based local feature extraction combined with embedding methods and temporal probabilistic framework is more promising; ii) since continuously accumulating data is necessary to achieve visual localization algorithms robust against appearance change, investigating indexing algorithms for a scalable visual localization algorithms is crucial in the scenarios of unbounded-growing database; iii) with the aim to the safety for autonomous driving, apart from accuracy, uncertainty in visual localization is a vital aspect, hence probabilistic framework should be explored further.

%
%
\bibliographystyle{splncs04}
\bibliography{mybib}

\begin{thebibliography}{10}
\providecommand{\url}[1]{\texttt{#1}}
\providecommand{\urlprefix}{URL }
\providecommand{\doi}[1]{https://doi.org/#1}

\bibitem{nonholonomicWiki}
\url{https://en.wikipedia.org/wiki/Nonholonomic\_system}

\bibitem{arandjelovic2016netvlad}
Arandjelovic, R., Gronat, P., Torii, A., Pajdla, T., Sivic, J.: Netvlad: Cnn
  architecture for weakly supervised place recognition. In: Proceedings of the
  IEEE conference on computer vision and pattern recognition. pp. 5297--5307
  (2016)

\bibitem{arandjelovic2012three}
Arandjelovic, R., Zisserman, A.: Three things everyone should know to improve
  object retrieval. In: CVPR (2012)

\bibitem{brachmann2017dsac}
Brachmann, E., Krull, A., Nowozin, S., Shotton, J., Michel, F., Gumhold, S.,
  Rother, C.: {DSAC}-differentiable {RANSAC} for camera localization. In: CVPR
  (2017)

\bibitem{brahmbhatt2018mapnet}
Brahmbhatt, S., Gu, J., Kim, K., Hays, J., Kautz, J.: Geometry-aware learning
  of maps for camera localization. In: CVPR (2018)

\bibitem{brubaker2013lost}
Brubaker, M.A., Geiger, A., Urtasun, R.: Lost! leveraging the crowd for
  probabilistic visual self-localization. In: CVPR (2013)

\bibitem{parra2019visual}
Bustos, A.P., Chin, T.J., Eriksson, A., Reid, I.: Visual slam: Why bundle
  adjust? In: ICRA (2019)

\bibitem{churchill2013experience}
Churchill, W., Newman, P.: Experience-based navigation for long-term
  localisation. The International Journal of Robotics Research  (2013)

\bibitem{do2015faemb}
Do, T.T., Tran, Q.D., Cheung, N.M.: Faemb: a function approximation-based
  embedding method for image retrieval. In: CVPR (2015)

\bibitem{Dzung18G2D}
Doan, A.D., Jawaid, A.M., Do, T.T., Chin, T.J.: {G2D}: from {GTA} to {Data}.
  arXiv preprint arXiv:1806.07381 pp.~1--9 (2018)

\bibitem{doan2019scalable}
Doan, A.D., Latif, Y., Chin, T.J., Liu, Y., Do, T.T., Reid, I.: Scalable place
  recognition under appearance change for autonomous driving. In: ICCV (2019)

\bibitem{he2016resnet}
He, K., Zhang, X., Ren, S., Sun, J.: Deep residual learning for image
  recognition. In: CVPR (2016)

\bibitem{jegou2012pcawhitening}
J{\'e}gou, H., Chum, O.: Negative evidences and co-occurences in image
  retrieval: The benefit of pca and whitening. In: ECCV (2012)

\bibitem{jegou2010vlad}
J{\'e}gou, H., Douze, M., Schmid, C., P{\'e}rez, P.: Aggregating local
  descriptors into a compact image representation. In: CVPR (2010)

\bibitem{jegou2014triangulation}
J{\'e}gou, H., Zisserman, A.: Triangulation embedding and democratic
  aggregation for image search. In: CVPR (2014)

\bibitem{junkins2009analytical}
Junkins, J.L., Schaub, H.: Analytical mechanics of space systems. American
  Institute of Aeronautics and Astronautics (2009)

\bibitem{kendall2016modelling}
Kendall, A., Cipolla, R.: Modelling uncertainty in deep learning for camera
  relocalization. In: ICRA (2016)

\bibitem{kendall2017geometric}
Kendall, A., Cipolla, R., et~al.: Geometric loss functions for camera pose
  regression with deep learning. In: CVPR (2017)

\bibitem{kendall2015posenet}
Kendall, A., Grimes, M., Cipolla, R.: Posenet: A convolutional network for
  real-time 6-dof camera relocalization. In: CVPR (2015)

\bibitem{ko2009gp}
Ko, J., Fox, D.: Gp-bayesfilters: Bayesian filtering using gaussian process
  prediction and observation models. Autonomous Robots  (2009)

\bibitem{krahenbuhl2018free}
Kr{\"a}henb{\"u}hl, P.: Free supervision from video games. In: CVPR (2018)

\bibitem{lepetit2009epnp}
Lepetit, V., Moreno-Noguer, F., Fua, P.: {EPnP}: An accurate o(n) solution to
  the {PnP} problem. IJCV  (2009)

\bibitem{OxfordRobotCar}
Maddern, W., Pascoe, G., Linegar, C., Newman, P.: 1 year, 1000 km: The oxford
  robotcar dataset. The International Journal of Robotics Research  (2017)

\bibitem{markley2007averaging}
Markley, F.L., Cheng, Y., Crassidis, J.L., Oshman, Y.: Averaging quaternions.
  Journal of Guidance, Control, and Dynamics  (2007)

\bibitem{menegatti2004image}
Menegatti, E., Zoccarato, M., Pagello, E., Ishiguro, H.: Image-based monte
  carlo localisation with omnidirectional images. Robotics and Autonomous
  Systems  (2004)

\bibitem{milford2012seqslam}
Milford, M.J., Wyeth, G.F.: Seqslam: Visual route-based navigation for sunny
  summer days and stormy winter nights. In: ICRA (2012)

\bibitem{murray2014generalized}
Murray, N., Perronnin, F.: Generalized max pooling. In: CVPR (2014)

\bibitem{richter2017playing}
Richter, S.R., Hayder, Z., Koltun, V.: Playing for benchmarks. In: ICCV (2017)

\bibitem{rubino2018practical}
Rubino, C., Del~Bue, A., Chin, T.J.: Practical motion segmentation for urban
  street view scenes. In: ICRA (2018)

\bibitem{sattler2017efficient}
Sattler, T., Leibe, B., Kobbelt, L.: Efficient \& effective prioritized
  matching for large-scale image-based localization  (2017)

\bibitem{sattler2018benchmarking}
Sattler, T., Maddern, W., Toft, C., Torii, A., Hammarstrand, L., Stenborg, E.,
  Safari, D., Okutomi, M., Pollefeys, M., Sivic, J., et~al.: Benchmarking
  {6DOF} outdoor visual localization in changing conditions. In: CVPR (2018)

\bibitem{schonberger2016structure}
Schonberger, J.L., Frahm, J.M.: Structure-from-motion revisited. In: CVPR
  (2016)

\bibitem{schonberger2018semantic}
Sch{\"o}nberger, J.L., Pollefeys, M., Geiger, A., Sattler, T.: Semantic visual
  localization. CVPR  (2018)

\bibitem{sunderhauf2013we}
S{\"u}nderhauf, N., Neubert, P., Protzel, P.: Are we there yet? challenging
  seqslam on a 3000 km journey across all four seasons. In: ICRA Workshop on
  Long-Term Autonomy (2013)

\bibitem{torii2015densevlad}
Torii, A., Arandjelovic, R., Sivic, J., Okutomi, M., Pajdla, T.: 24/7 place
  recognition by view synthesis. In: CVPR (2015)

\bibitem{tran2019device}
Tran, N.T., Le~Tan, D.K., Doan, A.D., Do, T.T., Bui, T.A., Tan, M., Cheung,
  N.M.: On-device scalable image-based localization via prioritized cascade
  search and fast one-many ransac. TIP  (2019)

\bibitem{tremblay2018training}
Tremblay, J., Prakash, A., Acuna, D., Brophy, M., Jampani, V., Anil, C., To,
  T., Cameracci, E., Boochoon, S., Birchfield, S.: Training deep networks with
  synthetic data: Bridging the reality gap by domain randomization. In: CVPR
  Workshop on Autonomous Driving (2018)

\bibitem{walch2017lstm}
Walch, F., Hazirbas, C., Leal-Taixe, L., Sattler, T., Hilsenbeck, S., Cremers,
  D.: Image-based localization using lstms for structured feature correlation.
  In: ICCV (2017)

\bibitem{apolloscape}
Wang, P., Huang, X., Cheng, X., Zhou, D., Geng, Q., Yang, R.: The {ApolloScape}
  open dataset for autonomous driving and its application. TPAMI  (2019)

\bibitem{whitley1994genetic}
Whitley, D.: A genetic algorithm tutorial. Statistics and computing  (1994)

\bibitem{wolf2002robust}
Wolf, J., Burgard, W., Burkhardt, H.: Robust vision-based localization for
  mobile robots using an image retrieval system based on invariant features.
  In: ICRA (2002)

\bibitem{wolf2005robust}
Wolf, J., Burgard, W., Burkhardt, H.: Robust vision-based localization by
  combining an image-retrieval system with monte carlo localization. IEEE
  Transactions on Robotics  (2005)

\bibitem{yu2010improved}
Yu, K., Zhang, T.: Improved local coordinate coding using local tangents. In:
  ICML (2010)

\end{thebibliography}
\end{document}